\documentclass{article}

\usepackage[autonum]{shortex}
\usepackage{xcolor}         % colors
\usepackage[sort&compress,numbers]{natbib}
\usepackage[nonatbib,final]{neurips_2022}

\usepackage[utf8]{inputenc} % allow utf-8 input
\usepackage[T1]{fontenc}    % use 8-bit T1 fonts
\usepackage{hyperref}       % hyperlinks
\usepackage{url}            % simple URL typesetting
\usepackage{booktabs}       % professional-quality tables
\usepackage{amsfonts}       % blackboard math symbols
\usepackage{nicefrac}       % compact symbols for 1/2, etc.
\usepackage{microtype}      % microtypography
\usepackage{xcolor}         % colors

\title{Bayesian Inference via Sparse Hamiltonian Flows}

\author{%
  Naitong Chen \qquad Zuheng Xu \qquad Trevor Campbell\\
  Department of Statistics\\
  University of British Columbia\\
  \texttt{[naitong.chen | zuheng.xu | trevor]@stat.ubc.ca} 
}

\begin{document}

\maketitle

\begin{abstract}
A Bayesian coreset is a small, weighted subset of data that replaces the full
dataset during Bayesian inference, with the goal of reducing computational
cost.  Although past work has shown empirically that there often
\emph{exists} a coreset with low inferential error, efficiently
constructing such a coreset remains a challenge.  Current methods tend to
be slow, require a secondary inference step after coreset construction, and
do not provide bounds on the data marginal evidence.  In this work, we
introduce a new method---\emph{sparse Hamiltonian flows}---that addresses
all three of these challenges.  The method involves first subsampling the data
uniformly, and then optimizing a Hamiltonian flow parametrized by coreset
weights and including periodic \emph{momentum quasi-refreshment steps}.
Theoretical results show that the method enables
an exponential compression of the dataset in a representative model,
and that the quasi-refreshment steps reduce the KL divergence to the target.
Real and synthetic experiments demonstrate that sparse Hamiltonian flows
provide accurate posterior approximations with significantly reduced runtime
compared with competing dynamical-system-based inference methods.
\end{abstract}

\section{Introduction}\label{sec:intro}

Bayesian inference 
provides a coherent approach to learning from data and uncertainty assessment
in a wide variety of complex statistical models. Two standard methodologies for
performing Bayesian inference in practice are 
Markov chain Monte Carlo (MCMC)
[\citealp{Robert04}; \citealp{Robert11}; \citealp[Ch.~11,12]{Gelman13}]
and variational inference (VI) \citep{Jordan99,Wainwright08}.
MCMC simulates a Markov chain that targets the posterior distribution.
In the increasingly common setting of large-scale data, most exact MCMC methods are 
intractable. This is essentially because simulating each MCMC step requires an (expensive) 
computation involving each data point, and 
many steps are required to obtain inferential results of a reasonable quality.
To reduce cost, a typical approach is to perform the computation for a random
subsample of the data, rather than the full dataset, at each step \cite{Bardenet15,Korattikara14,Maclaurin14,Welling11,Ahn12}
(see \citep{Quiroz18} for a recent survey).
However, recent work shows that the speed benefits are outweighed by the drawbacks;
uniformly subsampling at each step causes MCMC to either mix slowly or provide poor inferential
approximation quality \cite{Johndrow20,Nagapetyan17,Betancourt15,Quiroz18,Quiroz19}.
VI, on the other hand, posits a family
of approximations to the posterior and uses optimization to find the closest member, enabling
the use of scalable stochastic optimization algorithms \citep{Hoffmann13,Ranganath14}.
While past work involved simple parametric families,
recent work has developed flow families based on Markov chains \cite{Salimans15,Habib18}---and in particular, 
those based on Langevin and Hamiltonian dynamics \cite{Wolf16,Caterini18,Neal05,Geffner21,Zhang21,Thin21}.
However, because these Markov chains are typically designed to target the posterior distribution, 
each step again requires a computation involving all the data, making
KL minimization and sampling slow.
Repeated subsampling to reduce cost has the same issues that it does in MCMC.

Although repeated subsampling in each step 
of a Markov chain (for MCMC or VI) is not generally
helpful, recent work on \emph{Bayesian coresets}
\cite{Huggins16} has provided empirical evidence that
there often exists a \emph{fixed} small, weighted subset of the data---a
coreset---that one can use to replace the full dataset in a standard MCMC
or VI inference method \cite{Campbell19}.  
In order for the Bayesian coreset approach to be practically useful, one must (1) find a
suitable coreset that provides a good posterior
approximation; and (2) do so quickly enough that the speed-up of inference is worth
the time it takes to find the coreset.  
There is currently no option that satisfies these two desiderata. Importance
weighting methods \cite{Huggins16} are fast, but do not provide adequate
approximations in practice.  Sparse linear regression methods
\cite{Campbell19JMLR,Campbell18,Zhang21b} are fast and sometimes provide high-quality
approximations, but are very difficult to tune well.  And sparse variational
methods \cite{Campbell19,Manousakas20} find very high quality coreset
approximations without undue tuning effort, but are too slow to be practical.

This work introduces three key insights.
First, we can uniformly subsample the dataset once to pick the points in the coreset
(the weights still need to be optimized).  This selection is not only
significantly simpler than past algorithms; we show that it enables
constructing an \emph{exact} coreset---with KL divergence 0 to the
posterior---of size $O(\log_2(N))$ for $N$ data points in a representative
model (\cref{prop:subsample_basis}). 
Second, we can then construct a normalizing flow family based on 
Hamiltonian dynamics \cite{Rezende15,Caterini18,Neal05} 
that targets the coreset posterior (parametrized by coreset weights) rather than
the expensive full posterior.
This method address all
of the current challenges with coresets: it enables tractable \iid
sampling, provides a known density and normalization constant, and is tuned using straightforward KL
minimization with stochastic gradients.
It also addresses the inefficiency of Markov-chain-based VI families, as the Markov chain
steps are computed using the inexpensive coreset posterior density 
rather than the full posterior density.
The final insight is that past momentum tempering methods \cite{Caterini18} do not provide sufficient
flexibility for arbitrary approximation to the posterior, even in a simple setting (\cref{prop:problem}).
Thus, we introduce novel periodic \emph{momentum quasi-refreshment} steps that provably reduce the KL objective
(\cref{prop:quasi_marginal,prop:quasi_conditional}).
   The paper concludes with
real and synthetic experiments, demonstrating that sparse Hamiltonian flows
compare favourably to both current coreset compression methods and variational
flow-based families.  Proofs of all theoretical results may be found in the
appendix. 

It is worth noting that Hamiltonian flow posterior approximations based on a weighted data subsample were
also developed in concurrent work in the context of variational annealed importance sampling \cite{Jankowiak21},
and subsampling prior to weight optimization was developed in concurrent work on MCMC \cite{Naik22}.
In this work, we focus on incorporating Bayesian coresets into Hamiltonian-based normalizing flows to obtain fast 
and accurate posterior approximations.
\section{Background}\label{sec:background}

\subsection{Bayesian coresets}\label{sec:bkg-coresets}
We are given a target probability density $\pi(\theta)$ for variables $\theta \in \reals^d$
that takes the following form:
\[
\pi(\theta) = \frac{1}{Z} \exp\left(\sum_{n=1}^N f_n(\theta)\right) \pi_0(\theta).\label{eq:target}
\]
In a Bayesian inference problem with \iid data, 
$\pi_0$ is the prior density, the $f_n$ are the log-likelihood terms for $N$ data points,
and the normalization constant is in general not known.
The goal is to take samples from the distribution corresponding to density $\pi(\theta)$.

In order to avoid the $\Theta(N)$ cost of evaluating $\log\pi(\theta)$
or $\nabla\log \pi(\theta)$ (at least one of which must be 
conducted numerous times in most standard inference algorithms),
\emph{Bayesian coresets} \cite{Huggins16}
involve replacing the target with a surrogate density of the form
\[
\pi_w(\theta) = \frac{1}{Z(w)}\exp\left(\sum_{n=1}^N w_n f_n(\theta)\right) \pi_0(\theta),\label{eq:coreset}
\]
where $w\in\reals^N$, $w\geq 0$ are a set of weights. 
If $w$ has at most $M \ll N$ nonzeros, the $O(M)$ cost of evaluating $\log \pi_w(\theta)$ 
or $\nabla \log \pi_w(\theta)$
is a significant improvement upon the original $\Theta(N)$ cost.

The baseline method to construct a coreset is to draw a uniformly random
subsample of $M$ data points, and give each a weight of $N/M$; although this
method is fast in practice, it typically generates poor posterior
approximations. More advanced techniques generally involve significant user
tuning effort \cite{Huggins16,Campbell19JMLR,Campbell18,Zhang21b}. The current
state-of-the-art black box approach 
 formulates the problem as
variational inference \citep{Campbell19,Manousakas20} and provides a stochastic gradient
scheme using samples from $\pi_w$,
\[
w^\star = \argmin_{w\in\reals_+^N} \kl{\pi_w}{\pi}
\quad \text{s.t.}\quad\|w\|_0 \leq M.
\]
Empirically, this method tends to produce very high-quality coresets \cite{Campbell19}.
However, to estimate the gradient at each iteration of the optimization, we require MCMC samples from 
the weighted coreset posterior at that iteration. While generating MCMC samples 
from a sparse coreset posterior is not expensive, it is difficult to tune the algorithm to ensure the quality 
of these MCMC samples across iterations (as the weights, and thus the coreset posterior, change each iteration). 
The amount of tuning effort required makes the application 
of this method too slow to be practical. Once the coreset is constructed, all of the aforementioned methods 
require a secondary inference algorithm to take draws from $\pi_w$. 
Further, since $Z(w)$ is not known in general, it is
not tractable to use these methods to bound the marginal evidence $Z$.

\subsection{Hamiltonian dynamics}\label{sec:bkg-ham}
In this section we provide a very brief overview of some important 
aspects of a special case of Hamiltonian dynamics and its use in statistics; see \citep{Neal11}
for a more comprehensive overview.
The differential equation below in \cref{eq:ham_dynamics} describes
how a (deterministic) Hamiltonian system with position $\theta_t\in\reals^d$, momentum $\rho_t\in\reals^d$,
differentiable negative potential energy $\log\pi(\theta_t)$, and kinetic energy 
$\frac{1}{2} \rho_t^T\rho_t$ evolves over time $t\in\reals$:
\[
\der{\rho_t}{t} &= \nabla \log \pi(\theta_t) \qquad \der{\theta_t}{t} = \rho_t.\label{eq:ham_dynamics}
\]

For $t\in\reals$, define the mappings $H_t : \reals^{2d} \to \reals^{2d}$
that take $(\theta_s, \rho_s) \mapsto (\theta_{s+t}, \rho_{s+t})$ under the dynamics in \cref{eq:ham_dynamics}.
These mappings have two key properties that make Hamiltonian dynamics useful in statistics.
First, they are invertible, and preserve volume in the sense that $\left|\det \nabla H_t\right| = 1$.
In other words, they provide tractable density transformations: 
for any density $q$ on $\reals^{2d}$ and pushforward $q_t$ on $\reals^{2d}$ under the mapping $H_t$,
we have that $q_t(\cdot, \cdot) = q\left(H_t^{-1}(\cdot, \cdot)\right)$.
Second, the
\emph{augmented target density} $\bpi(\theta, \rho)$ on $\reals^{2d}$
corresponding to independent draws from $\pi$ and $\distNorm(0,I)$,
\[
\bpi(\theta, \rho)  \propto \pi(\theta)\cdot\exp\left(-\frac{1}{2}\rho^T\rho\right), \label{eq:bpi}
\]
is invariant under the mappings $H_t$, i.e.,
$\bpi(H_t(\cdot, \cdot)) = \bpi(\cdot,\cdot)$.
Given these properties, Hamiltonian Monte Carlo \cite{Neal11,Neal96} constructs
a Gibbs sampler for $\bpi$ that interleaves Hamiltonian dynamics with periodic stochastic momentum refreshments $\rho\distas\distNorm(0, I)$. 
Upon completion, the $\rho$ component of the samples can be dropped to
obtain samples from the desired target $\pi$.

In practice, one approximately simulates the dynamics in \cref{eq:ham_dynamics}
using the leapfrog method, which 
involves interleaving three discrete transformations with step size $\epsilon >0$,
\[
\hrho_{k+1} &= \rho_k + \frac{\epsilon}{2}\nabla \log\pi(\theta_k) &
\theta_{k+1} &= \theta_{k} + \epsilon \hrho_{k+1} &
\rho_{k+1} &= \hrho_{k+1} + \frac{\epsilon}{2}\nabla \log\pi(\theta_{k+1}).
\label{eq:leapfrog}
\]
Denote the map constructed by applying these three steps
in sequence $T_{\epsilon} : \reals^{2d}\to\reals^{2d}$.  As the transformations
in \cref{eq:leapfrog} are all shear, $T_{\epsilon}$ is also
volume-preserving, and for small enough step size $\epsilon$ it nearly
maintains the target invariance.
Note also that evaluating a single application of $T_{\epsilon}$ 
is of $O(Nd)$ complexity, which is generally expensive in the large-data (large-$N$) regime.

\subsection{VI via Hamiltonian dynamics}\label{sec:bkg-vihmc}

Since the mapping $T_{\epsilon}$ is invertible and volume-preserving,
it is possible to tractably compute the density of the pushforward of a 
reference distribution $q(\cdot, \cdot)$ under repeated applications of it.
In addition, this repated application of $T_{\epsilon}$ resembles the steps of Hamiltonian Monte Carlo (HMC) 
\cite{Neal11}, which we know converges in distribution to the target posterior distribution. 
\citep{Caterini18,Neal05} use these facts to construct a normalizing
flow \cite{Rezende15} VI family. 
However, there are two issues with this methodology.
First, the $O(Nd)$ complexity of evaluating each step $T_{\epsilon}$ 
makes training and simulating from this flow computationally expensive.
Second, Hamiltonian dynamics on its own creates a flow with insufficient flexibility to match a target
$\bpi$ of interest. In particular, 
given a  density $q(\cdot,\cdot)$ and pushforward $q_t(\cdot, \cdot)$ under $H_t$,
we have
\[
\forall t\in\reals, \quad \kl{q_t}{\bpi} = \kl{q}{\bpi}. \label{eq:constantkl}
\]
In other words, Hamiltonian dynamics itself cannot reduce the KL divergence to $\bpi$; it 
simply interchanges potential and kinetic energy.
\citep{Caterini18} address this issue by instead deriving their flow from
\emph{tempered} Hamiltonian dynamics: for an integrable tempering 
function $\gamma : \reals \to \reals$,
\[
\der{\rho_t}{t} &= \nabla \log \pi(\theta_t) - \gamma(t)\rho_t \qquad \der{\theta_t}{t} = \rho_t.\label{eq:tempered_ham_dynamics}
\]
The discretized version of the dynamics in \cref{eq:tempered_ham_dynamics} corresponds
to multiplying the momentum by a tempering value $\alpha_k > 0$ after the
$k^\text{th}$ application of $T_{\epsilon}$. By scaling the momentum, one provides
the normalizing flow with the flexibility to change the kinetic energy at each step.
However, we show later in \cref{prop:problem} that just tempering the momentum does
not provide the required flow flexibility, even for a simple representative Gaussian target.

A related line of work uses the mapping $T_{\epsilon}$ for variational annealed
importance sampling \cite{Geffner21,Zhang21,Thin21}. The major difference between these
methods and the normalizing flow-based methods is that the auxiliary variable is 
(partially) stochastically refreshed via $\rho \distas \distNorm(0, I)$ after
applications of $T_{\epsilon}$. One is then forced to minimize the KL divergence
between the joint distribution of $\theta$ and all of the auxiliary momentum variables
under the variational and augmented target distributions.

\section{Sparse Hamiltonian flows}\label{sec:sparse_flows}
In this section we present sparse Hamiltonian flows,
a new method to construct and draw samples from
Bayesian coreset posterior approximations.  
We first present a method and supporting theory for
selecting the data points to be included in the coreset,
then discuss building a sparse flow with these points,
and finally introduce quasi-refreshment steps to give the flow
family enough flexibility to match the target distribution.
Sparse Hamiltonian flows enables
tractable \iid sampling, provides a tractable density and 
normalization constant, and is
constructed by minimizing the KL divergence to the posterior with
simple stochastic gradient estimates. 

\subsection{Selection via subsampling}\label{sec:select-sample}
The first step in our algorithm is to choose a uniformly
random subsample of $M$ points from the full dataset;
these will be the data points that comprise the coreset. Without loss of generality, 
we assume these are the first $M$ points.
The key insight in this work is that while 
subsampling with importance weighting does 
not typically provide good coreset approximations \cite{Huggins16},
a uniformly random subset of the $N$ log-likelihood potential
functions $\{f_{1}, \dots, f_M\}$ still provides
a good \emph{basis} for approximation with high probability. 
\cref{prop:subsample_basis} provides the precise statement
of this result for a representative example model \cref{eq:simple_model}.
In particular, \cref{prop:subsample_basis} asserts that 
as long as we set our coreset size $M$ to be proportional to 
$d \log_2 N$, the optimal coreset posterior approximation
will be \emph{exact}, i.e., have 0 KL divergence to the true posterior, with probability at least $1- N^{-\frac{d}{2}}(\log_2 N)^{\frac{d}{2}}$.
Thus we achieve an exponential compression of the dataset, $N \to \log_2 N$, without losing any fidelity.
Note that we will still need a method to choose the weights $w_1, \dots, w_M$
for the $M$ points, but the use of uniform selection
rather than a one-at-a-time approach  
\cite{Campbell18,Campbell19JMLR,Campbell19} substantially simplifies
the construction.
In \cref{prop:subsample_basis}, $C$ is the universal constant from \citep[Corollary 1.2]{Boroczky03},
which provides an upper bound on the number of spherical balls of some fixed radius needed to cover a 
$d$-dimensional unit sphere.
\bprop \label{prop:subsample_basis}
Consider a Bayesian Gaussian location model:
\[
	\theta &\distas \distNorm(0, I) \quad\text{and}\quad
	\forall n\in[N], \quad X_n \distiid \distNorm(\theta, I),
	\label{eq:simple_model}
\]
where $\theta, X_n\in\reals^d$ for $d\in\nats$.
Suppose the true data generating parameter $\theta = 0$,
and set $M = \log_2(A_d N^d(\log N)^{-d/2}) + C$ 
where $A_d = e^{\frac{d}{2}}d^{\frac{3}{2}}\log(1+d)$.
Then the optimal coreset $\pi_{w^\star}$ for the model \cref{eq:simple_model} 
built using a uniform
subsample of data of size $M$ satisfies
\[
\limsup_{N\to\infty} \frac{\Pr\left(\kl{\pi_{w^\star}}{\pi} \neq 0\right)}{N^{-\frac{d}{2}}(\log N)^{\frac{d}{2}}} \leq 1.
\]
\eprop

\subsection{Sparse flows}
Upon taking a uniform subsample of $M$ data points 
from the full dataset, 
we consider the sparsified Hamiltonian dynamics
initialized at $\theta_0, \rho_0 \distas q(\cdot, \cdot)$ for reference density\footnote{The reference $q$ can also have its own variational parameters
to optimize, but in this paper we leave it fixed.} $q(\cdot,\cdot)$, 
\[
\der{\rho_t}{t} &= \nabla \log \pi_w(\theta_t) \qquad \der{\theta_t}{t} = \rho_t.\label{eq:wham_dynamics}
\]
Much like the original Hamiltonian dynamics for the full target density,
the sparsified Hamiltonian dynamics \cref{eq:wham_dynamics} targets
the augmented coreset posterior with density
$\bpi_w(\theta, \rho)$ on $\reals^{2d}$,
\[
\bpi_w(\theta,\rho) \propto \pi_w(\theta) \exp\left(-\frac{1}{2}\rho^T\rho\right).
\]
Discretizing these dynamics yields a leapfrog 
method similar to \cref{eq:leapfrog}
with three interleaved steps,
\[
\hrho_{k+1} &= \rho_k + \frac{\epsilon}{2}\nabla \log\pi_w(\theta_k) &
\theta_{k+1} &= \theta_{k} + \epsilon \hrho_{k+1} &
\rho_{k+1} &= \hrho_{k+1} + \frac{\epsilon}{2}\nabla \log\pi_w(\theta_{k+1}).
\label{eq:wleapfrog}
\]
Denote the map constructed by applying these three steps in sequence $T_{w,\epsilon} : \reals^{2d}\to\reals^{2d}$.
Like the original leapfrog method, these transformations are both invertible and shear, and thus preserve volume;
and for small enough step size $\epsilon$, they approximately maintain the invariance of $\bpi_w(\theta, \rho)$.
However, since $w$ only has the first $M$ entries nonzero,
\[
	\nabla \log \pi_w(\theta_k) &= \sum_{m=1}^M w_m \nabla \log f_m(\theta_k),
\]
and thus a coreset leapfrog step can be taken in $O(Md)$ time, as opposed to $O(Nd)$ 
time in the original approach. Given that \cref{prop:subsample_basis} recommends
setting $M \approx d\log_2(N)$, we have achieved an exponential reduction
in computational cost of running the flow.

However, as before, the weighted sparse leapfrog flow is not sufficient on its own
to provide a flexible variational family. In particular,
we know that $T_{w,\epsilon}$
nearly maintains the distribution $\bpi_w$ as invariant.
We therefore need a way to modify the distribution
of the momentum variable $\rho$.
One option is to include
a tempering of the form \cref{eq:tempered_ham_dynamics}
into the sparse flow.
However, \cref{prop:problem} shows
that even \emph{optimal} tempering does not provide the flexibility
to match a simple Gaussian target $\bpi$.
\bprop\label{prop:problem}
Let $\theta_t, \rho_t \in \reals$ 
follow the tempered Hamiltonian dynamics \cref{eq:tempered_ham_dynamics}
targeting $\pi = \distNorm(0, \sigma^{2})$, $\sigma > 0$,
with initial distribution $\theta_0 \distas\distNorm(\mu, 1)$, $\rho_0 \distas\distNorm(0, \beta^2)$
for initial center $\mu\in\reals$ and momentum scale $\beta>0$.
Let $q_t$ be the distribution of $(\theta_t, \rho_t)$.
Then
\[
\inf_{t>0, \beta > 0, \gamma : \reals_+\to\reals} \kl{q_t}{\bpi} &\geq \log\frac{1+\mu^2}{4\sigma}.
\]
Note that if $\gamma(t) = 0$ identically, 
then $\forall t\geq 0$, $\kl{q_t}{\bpi} = \kl{q_0}{\bpi}$.
\eprop
The intuition behind \cref{prop:problem} is that while adding a tempering $\gamma(t)$
enables one to change the total energy by scaling the momentum,
it does not allow one fine enough control on the distribution of the momentum.
For example, if $\EE[\rho] \neq 0$
under the current flow approximation, we cannot scale the momentum to force $\EE[\rho]=0$ as it should be
under the augmented target; intuitively, we also need the ability to shift or recenter the momentum as well.

\begin{minipage}[t]{0.45\textwidth}
\begin{algorithm}[H]
\caption{\texttt{SparseHamFlow}}\label{alg:sampleshf}
\begin{algorithmic}
\Require $\theta_0$, $\rho_0$, $w$, $\epsilon$, $\lambda$, $L$, $R$
\State $J \gets 0$, and $(\theta, \rho)\gets(\theta_0, \rho_0)$
\For{$r=1, \dots, R$}
\For{$\ell=1,\dots,L$}
\State Sparse flow leapfrog:
\State $\theta, \rho \gets T_{w, \epsilon}(\theta, \rho)$
\EndFor
\State Accumulate log Jacobian determinant:
\State $J \gets J +\log\abs{\det \frac{\partial R_{\lambda_{r}}}{\partial \rho}(\rho, \theta)}$
\State Quasi-refreshment:
\State $\rho \gets R_{\lambda_{r}}(\rho, \theta)$
\EndFor
\State \Return $\theta, \rho, J$
\end{algorithmic}
\end{algorithm}
\end{minipage}
\hfill
\begin{minipage}[t]{0.52\textwidth}
\begin{algorithm}[H]
\caption{\texttt{Estimate\_ELBO}}\label{alg:estelbo}
\begin{algorithmic}
\Require $q$, $\pi_0$, $w$, $\epsilon$, $\lambda$, $L$, $R$, $S$
\State $(\theta_0, \rho_0) \distas q(\cdot, \cdot)$
\State Forward pass:
\State $\theta, \rho, J \gets$ \texttt{SparseHamFlow}$ \left(\theta_0, \rho_0, w,\epsilon, \lambda, L, R\right)$
\State Obtain unbiased ELBO estimate:
\State $(n_1, \dots, n_S) \distiid \distUnif(\{1,2,\dots,N\})$
\State $\log\bar{p} \gets \log \pi_0(\theta) + \frac{N}{S}\sum_{s=1}^S f_{n_s}(\theta) +$ 
\State $\quad\quad\quad\quad\log\distNorm(\rho \given 0, I)$
\State $\log\bar{q} \gets q(\theta_0, \rho_0) - J$
\State \Return $\log\bar{p} - \log\bar{q}$
\end{algorithmic}
\end{algorithm}
\end{minipage}

\subsection{Quasi-refreshment}\label{sec:quasi-refresh}
Rather than resampling the momentum variable from its target
marginal---which removes the ability to evaluate the density 
of $\theta_t,\rho_t$---in this work we introduce deterministic \emph{quasi-refreshment}
moves that enable the flow to strategically update the momentum without losing
the ability to compute the density and normalization constant of $\theta_t,
\rho_t$ (i.e., we construct a normalizing flow \cite{Rezende15}).
Here we introduce the notion of \emph{marginal} quasi-refreshment, which tries to make the marginal distribution of $\rho_t$
match the corresponding marginal distribution of the augmented target $\bpi_w$.
\cref{prop:quasi_marginal} shows that  marginal quasi-refreshment is guaranteed to reduce the KL divergence.
\bprop\label{prop:quasi_marginal}
Consider the state $\theta_t, \rho_t\in\reals^d$ of the flow at step $t$, 
and the augmented target distribution $\theta, \rho \distas \bpi$.
Suppose that we have a bijection 
$R : \reals^d \to \reals^d$ such that
$R(\rho_t) \eqd \rho$. Then
\[
\kl{\theta_t, R(\rho_t)}{\theta,\rho} &=\kl{\theta_t,\rho_t}{\theta,\rho} - \kl{\rho_t}{\rho}.
\]
\eprop

\begin{wrapfigure}{r}{0.5\textwidth}
	\centering
	\includegraphics[width = 0.5\textwidth]{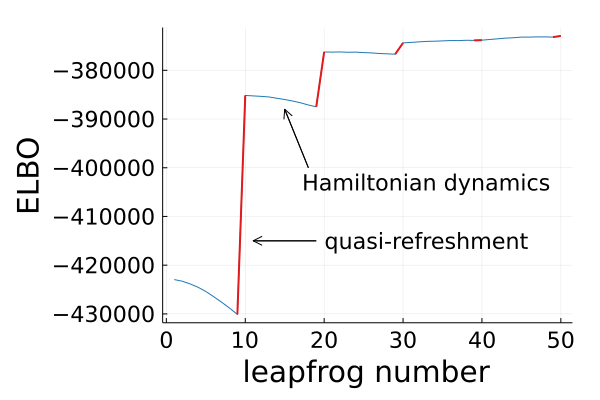}
	\caption{ELBO across leapfrog steps.}
	\label{fig:step}
\end{wrapfigure}

See \cref{sec:quasi-appendix} for the proof of \cref{prop:quasi_marginal} and a general treatment 
of quasi-refreshment; for simplicity, we focus on the type of quasi-refreshment that we use
in the experiments. In particular, if we are willing to make an assumption
about the marginal distribution of
$\rho_t$ at step $t$ of the flow, we can introduce a tunable family of functions $R_\lambda$ 
with parameters $\lambda$ that is flexible enough to set $R_\lambda(\rho_t) \eqd \rho$ for some $\lambda$, 
and include optimization of $\lambda$ along with the coreset weights. It is important to note that
this assumption on the distribution of $\rho_t$ is not related to the posterior $\pi$.
As an example, in this work we assume that $\rho_t \distas \distNorm(\mu, \Lambda^{-1})$ 
for some unknown mean $\mu$ and diagonal precision $\Lambda$, which enables us to simply set
\[
R_\lambda(x) = \Lambda\left(x - \mu\right). \label{eq:marginal_standardization}
\]
We then include $\lambda = (\mu,\Lambda)$ as parameters
to be optimized along with the coreset weights $w$ (each quasi-refreshment step will have its own 
set of parameters $\mu,\Lambda$).
Even when this assumption does not hold exactly,
the resulting form of \cref{eq:marginal_standardization} enables
the refreshment step to both shift and scale (i.e., standardize) the momentum as desired,
and is natural to implement as part of a single optimization routine.

\cref{fig:step} provides an example of the effect of quasi-refreshment
in a synthetic Gaussian location
model (see \cref{sec:synthetic} for details). In particular, it shows the 
evidence lower bound (ELBO) as a function of leapfrog step number in a trained sparse
Hamiltonian flow with the
quasi-refreshment scheme in \cref{eq:marginal_standardization}. 
While the estimated ELBO values stay relatively stable
across leapfrog steps in between quasi-refreshments (as expected by \cref{eq:constantkl}), 
the quasi-refreshment steps (colored red) cause the ELBO to increase drastically. 
We note that the ELBO does not stay exactly constant because the Hamiltonian 
dynamics targets the coreset posterior instead of the true posterior, and is simulated 
approximately using leapfrog steps. As the series of transformations brings the approximated 
density closer to the target, the quasi-refreshment steps no longer change the ELBO much, 
signalling the convergence of the flow's approximation of the target. It is thus clear that the
marginal quasi-refreshments indeed decrease the KL, as shown in
\cref{prop:quasi_marginal}.

\subsection{Algorithm}

In this section, we describe the procedure for training and
generating samples from a sparse Hamiltonian flow. 
As a normalizing flow, a sparse Hamiltonian flow can be trained 
by maximizing the augmented ELBO 
using usual stochastic gradient methods (e.g.~as in \cite{Rezende15}), where the transformations follow \cref{eq:wleapfrog}
with a periodic quasi-refreshment. Here and in the experiments we
focus on the shift-and-scale quasi-refreshment in \cref{eq:marginal_standardization}. 

We begin by selecting a subset of $M$
data points chosen uniformly randomly from the full data.
Next we select
a total number $R$ of quasi-refreshment steps,
and a number $L$ of leapfrog steps between
each quasi-refreshment.
The flow parameters to be optimized consist of
the quasi-refreshment parameters $\lambda = (\lambda_r)_{r=1}^R$,
the $M$ coreset weights $w = (w_m)_{m=1}^M$,
and the leapfrog step sizes $\epsilon = (\epsilon_i)_{i=1}^{d}$; note
that we use a separate step size $\epsilon_i$ per latent variable dimension $i$
in \cref{eq:wleapfrog} \citep[Sec.~4.2]{Neal11}. This modification enables the flow
to fit nonisotropic target distributions.

We initialize the weights to $N/M$ (i.e., a uniform coreset),
and select an initial step size for all dimensions.
We use a warm start to initialize
the parameters $\lambda_r = (\mu_r, \Lambda_r)$ of the quasi-refreshments. 
Specifically, using the initial leapfrog step sizes and coreset weights, we pass a batch of
samples from the reference density $q(\cdot,\cdot)$ through the flow
up to the first quasi-refreshment step. We initialize $\mu_1, \Lambda_1$ 
to the empirical mean and diagonal precision of the samples at that point.
We then apply the initialized first quasi-refreshment to the momentum, 
proceed with the second sequence of leapfrog steps, and repeat
until we have initialized all quasi-refreshments $r=1,\dots, R$.

Once the parameters are initialized, 
we log-transform the step sizes, weights, and quasi-refreshment diagonal scaling matrices
to make them unconstrained during optimization.
We obtain an unbiased estimate
of the augmented ELBO gradient by
applying automatic differentiation \citep{Baydin18,Kucukelbir17} 
to the ELBO estimation function \cref{alg:estelbo},
and optimize all parameters jointly using
a gradient-based stochastic optimization technique such as SGD
\citep{robbins1951stochastic,Bottou2004} and ADAM \citep{Kingma14}. 
 Once trained, we can obtain samples from
the flow via \cref{alg:sampleshf}. 
\section{Experiments}\label{sec:experiments}
In this section, we compare our method against other Hamiltonian-based VI methods 
and Bayesian coreset construction methods. Specifically, 
we compare the quality of posterior approximation, as well as
the training and sampling times of sparse Hamiltonian flows (\texttt{SHF}),
Hamiltonian importance sampling (\texttt{HIS}) \cite{Caterini18}, and
unadjusted Hamiltonian annealing (\texttt{UHA}) \cite{Geffner21} using real and
synthetic datasets. We compare with two variants of \texttt{HIS} and \texttt{UHA}:
``\texttt{-Full},'' in which we train using in-flow minibatching as suggested by
\citep{Caterini18,Geffner21}, but compute evaluation metrics using the full-data flow;
and ``\texttt{-Coreset},'' in which we base the flow on a uniformly subsampled coreset.
We also include sampling times of adaptive HMC and NUTS
\citep[Alg.~5 and~6]{Hoffman14} using the full dataset.
Finally, we compare the
quality of coresets constructed by \texttt{SHF}
to those obtained using uniform subsampling
(\texttt{UNI}) and Hilbert coresets with
orthogonal matching pursuit (\texttt{Hilbert-OMP})
\citep{Campbell19JMLR,pati1993orthogonal}. All experiments are performed on a machine with an 
Intel Core i7-12700H processor and 32GB memory. Code is available
at \url{https://github.com/NaitongChen/Sparse-Hamiltonian-Flows}. 
Details of the experiments are in \cref{supp:expt}.

\subsection{Synthetic Gaussian}\label{sec:synthetic}

We first demonstrate the performance of \texttt{SHF} on a 
synthetic Gaussian-location model,
\[
	\theta &\distas \distNorm(0, I) \quad\text{and}\quad
	\forall n\in[N], \quad X_n \distiid \distNorm(\theta, cI),
\]
where $\theta, X_n\in\reals^d$. We set $c = 100$, $d = 10$, $N = 10,000$.
This model
has a closed from posterior distribution 
$\pi =\distNorm\left(\frac{\sum_{n=1}^N X_n}{c+N}, \frac{c}{c+N}I\right)$. More
details may be found in \cref{supp:syndetail}.

\begin{figure*}[t!]
    \centering 
\begin{subfigure}[b]{.32\textwidth} 
    \scalebox{1}{\includegraphics[width=\textwidth]{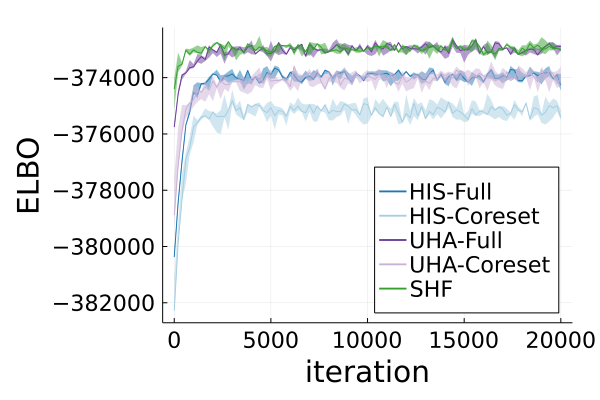}}
    \caption{(a)\label{fig:syngaussianelbo}}
\end{subfigure}
\hfill
\centering
\begin{subfigure}[b]{0.32\textwidth}
    \scalebox{1}{\includegraphics[width=\textwidth]{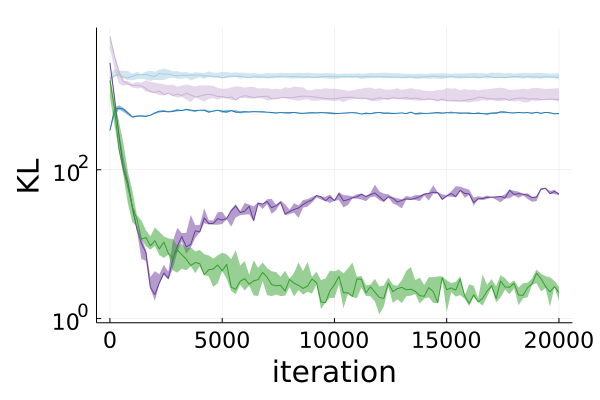}}
    \caption{(b)\label{fig:syngaussiankl}}
\end{subfigure}
\hfill
\centering 
\begin{subfigure}[b]{.32\textwidth} 
    \scalebox{1}{\includegraphics[width=\textwidth]{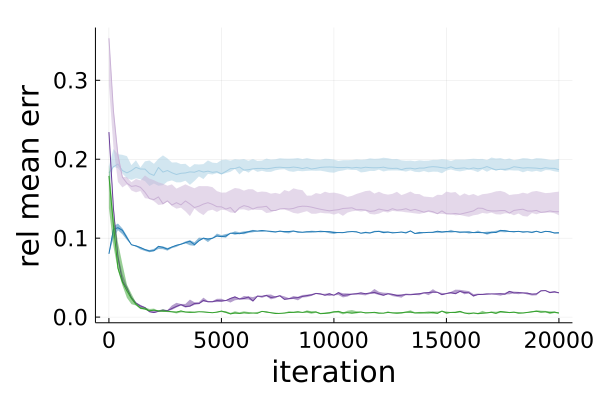}}
    \caption{(c)\label{fig:syngaussianmean}}
\end{subfigure}
\hfill
\centering
\begin{subfigure}[b]{0.32\textwidth}
    \scalebox{1}{\includegraphics[width=\textwidth]{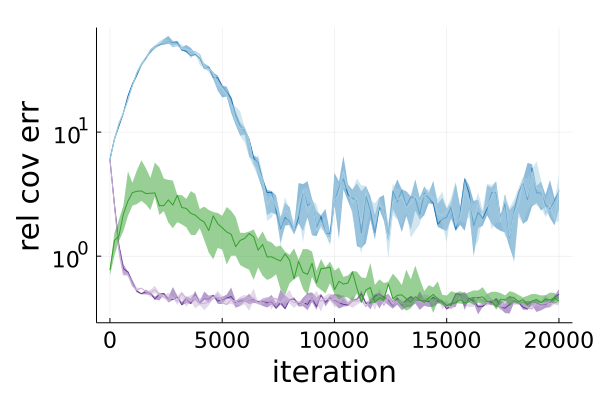}}
    \caption{(d)\label{fig:syngaussiancov}}
\end{subfigure}
\hfill
\centering
\begin{subfigure}[b]{0.32\textwidth}
    \scalebox{1}{\includegraphics[width=\textwidth]{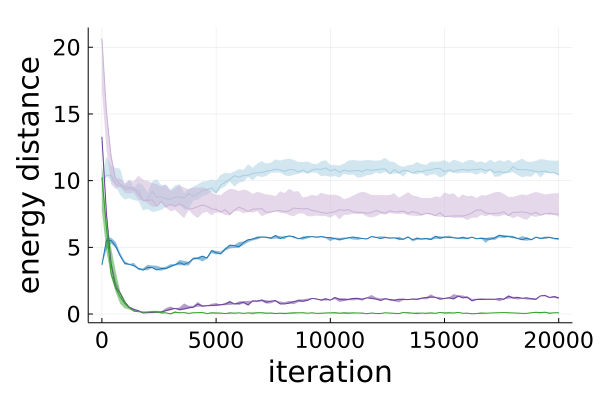}}
    \caption{(e)\label{fig:syngaussianed}}
\end{subfigure}
\hfill
\centering
\begin{subfigure}[b]{0.32\textwidth}
    \scalebox{1}{\includegraphics[width=\textwidth]{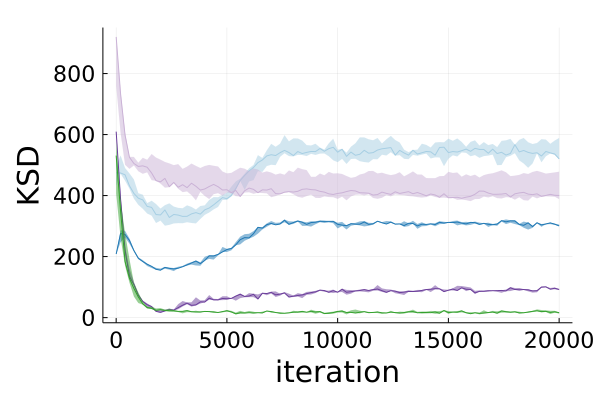}}
    \caption{(f)\label{fig:syngaussianksd}}
\end{subfigure}
\hfill

\caption{ELBO (\cref{fig:syngaussianelbo}), KL divergence (\cref{fig:syngaussiankl}), 
relative 2-norm mean error (\cref{fig:syngaussianmean}), relative Frobenius norm covariance error (\cref{fig:syngaussiancov}), energy distance (\cref{fig:syngaussianed}), and IMQ KSD \cite{gorham2017measuring} (\cref{fig:syngaussianksd}) for synthetic Gaussian. The lines indicate the median, and error regions indicate 25$^\text{th}$ to 75$^\text{th}$ percentile from 5 runs.}
\label{fig:synmetric}
\end{figure*}

\cref{fig:syngaussianelbo} compares the ELBO values of \texttt{SHF},
\texttt{HIS}, and \texttt{UHA} across all optimization iterations. We can see that 
\texttt{SHF} and \texttt{UHA-Full}
result in the highest ELBO, and hence tightest bound on
the log normalization constant of the target.
In this problem, since we have access to the exact posterior distribution in closed form,
we can also estimate the $\theta$-marginal KL divergence directly, as shown in \cref{fig:syngaussiankl}.  
Here we see the posterior approximation produced by
\texttt{SHF} provides a significantly lower KL than the other competing methods. 
\cref{fig:syngaussianmean,fig:syngaussiancov,fig:syngaussianed,fig:syngaussianksd} show, through 
a number of other metrics, that \texttt{SHF} provides a higher quality posterior approximation than others.
It is worth noting that while the relative covariance error for \texttt{SHF} takes long to converge, we observe 
a monotonic downward trend in both the relative mean error and KL divergence of \texttt{SHF}. This means 
that for this particular problem, our method finds the centre of the target before fine tuning the covariance. 
We also note that a number of metrics go up for \texttt{UHA-full} because it operates on the augmented space 
based on a sequence of distributions that bridge some simple initial distribution and the target distribution. 
Therefore, it is not guaranteed that all steps of optimization improve the quality of approximation on the 
marginal space of the latent variables of interest, which is what the plots in \cref{fig:synmetric} show.
%, with a roughly equally accurate approximate posterior 
%covariance as in \texttt{UHA-Full}.

\captionsetup[subfigure]{labelformat=empty}

\captionsetup[subfigure]{labelformat=empty}
\begin{figure*}[t!]
    \centering 
\begin{subfigure}[b]{.32\textwidth} 
    \scalebox{1}{\includegraphics[width=\textwidth]{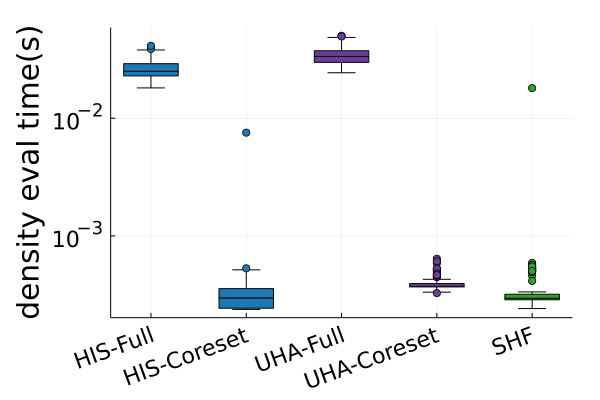}}
    \caption{(a)\label{fig:syngaussiandensity}}
\end{subfigure}
\hfill
\centering
\begin{subfigure}[b]{0.32\textwidth}
    \scalebox{1}{\includegraphics[width=\textwidth]{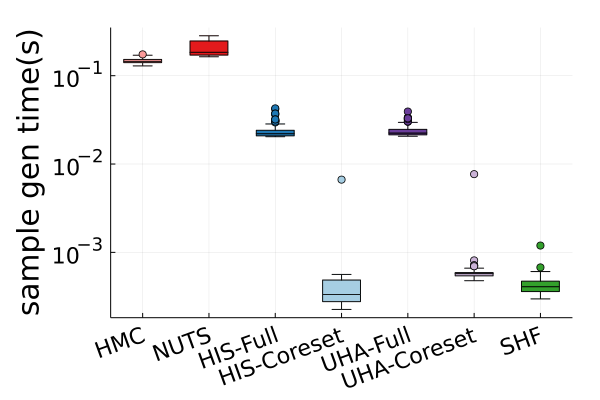}}
    \caption{(b)\label{fig:syngaussiansample}}
\end{subfigure}
\hfill
\centering
\begin{subfigure}[b]{0.32\textwidth}
    \scalebox{1}{\includegraphics[width=\textwidth]{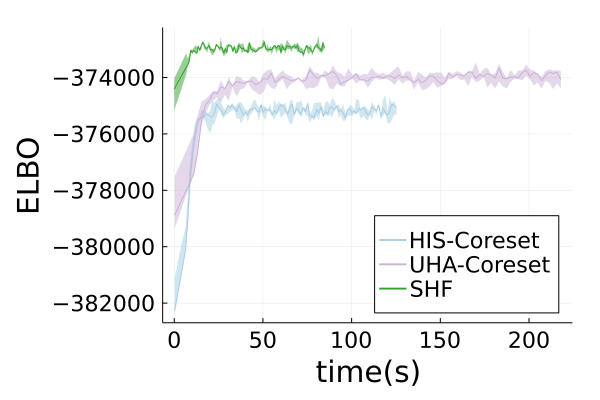}}
    \caption{(c)\label{fig:syngaussianelbotime}}
\end{subfigure}

\caption{Density evaluation (\cref{fig:syngaussiandensity}) and sample generation time (\cref{fig:syngaussiansample}) (100 samples), and ELBO versus time during training (\cref{fig:syngaussianelbotime}) for synthetic Gaussian. The lines indicate the median, and error regions indicate 25$^\text{th}$ to 75$^\text{th}$ percentile from 5 runs.}
\label{fig:syntime}
\end{figure*}
\captionsetup[subfigure]{labelformat=empty}
\begin{figure*}[t!]
    \centering 
\begin{subfigure}[b]{.32\textwidth} 
    \scalebox{1}{\includegraphics[width=\textwidth]{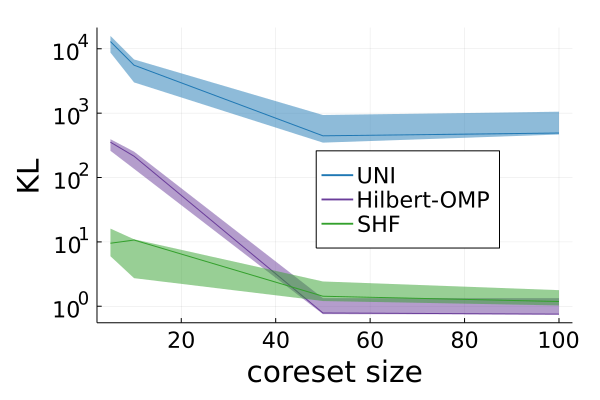}}
    \caption{(a)\label{fig:syncoresetkl}}
\end{subfigure}
\hfill
\centering
\begin{subfigure}[b]{0.32\textwidth}
    \scalebox{1}{\includegraphics[width=\textwidth]{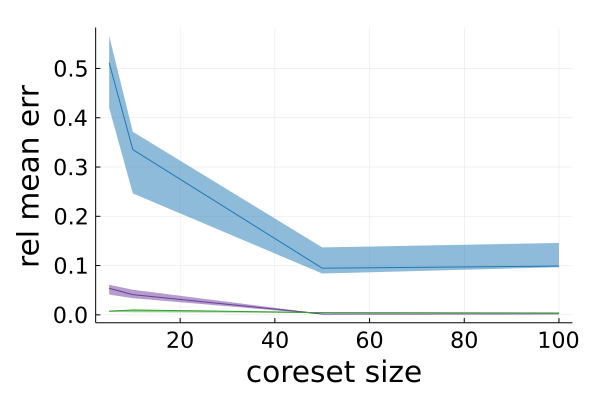}}
    \caption{(b)\label{fig:syncoresetmean}}
\end{subfigure}
\hfill
\centering
\begin{subfigure}[b]{0.32\textwidth}
    \scalebox{1}{\includegraphics[width=\textwidth]{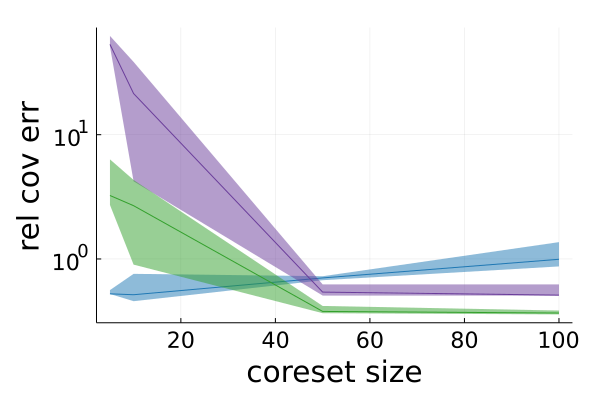}}
    \caption{(c)\label{fig:syncoresetcov}}
\end{subfigure}
\hfill
\begin{subfigure}[b]{0.32\textwidth}
    \scalebox{1}{\includegraphics[width=\textwidth]{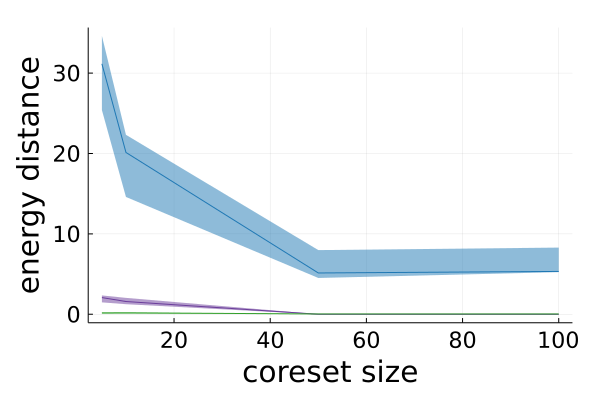}}
    \caption{(d)\label{fig:syncoreseted}}
\end{subfigure}
\begin{subfigure}[b]{0.32\textwidth}
    \scalebox{1}{\includegraphics[width=\textwidth]{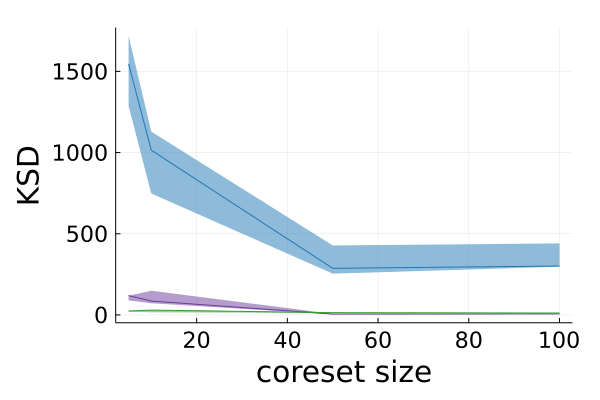}}
    \caption{(e)\label{fig:syncoresetksd}}
\end{subfigure}
\caption{Estimated KL divergence (\cref{fig:syncoresetkl}),
relative 2-norm mean error (\cref{fig:syncoresetmean}),
relative Frobenius norm covariance error (\cref{fig:syncoresetcov}), 
energy distance (\cref{fig:syncoreseted}), and IMQ KSD (\cref{fig:syncoresetksd}) versus coreset size. 
The lines indicate the median, and error regions indicate 25$^\text{th}$ to 75$^\text{th}$ percentile from 5 runs.}
\label{fig:syncoreset}
\end{figure*}

\cref{fig:syngaussiandensity,fig:syngaussiansample} show the time required for
each method to 
evaluate the density of the joint distribution of $\theta, \rho$ and to generate samples.
It is clear that the use of a coreset improves the
density evaluation and sample generation time by more than an order of magnitude. 
\cref{fig:syngaussianelbotime} compares the training times of \texttt{SHF},
\texttt{HIS-Coreset}, and \texttt{UHA-Coreset} (recall that due to the use of 
subsampled minibatch flow dynamics,
\texttt{HIS-Full} and \texttt{UHA-Full} share the same training time as their \texttt{-Coreset} versions). 
The relative training speeds generally match those of sample generation from the target posterior. 
%\texttt{HIS} is slower than \texttt{SFH}
%since it requires a random resampling of the data points in each iteration,
%while \texttt{SHF} picks a fixed coreset basis and only optimizes the
%corresponding weights. \texttt{UHA} takes the longest per iteration due to the use of bridging
%distributions and auxiliary momentum variables, which result in more complicated
%gradient expressions (although the ELBO appears to converge more quickly).

Finally, \cref{fig:syncoreset} compares the quality of coresets
constructed via \texttt{SHF}, uniform subsampling (\texttt{UNI}),
and Hilbert coresets with orthogonal matching pursuit (\texttt{Hilbert-OMP}).
Note that in this problem, the Laplace approximation is exact (the true posterior is Gaussian),
and hence \texttt{Hilbert-OMP} constructs a coreset using samples from the true posterior.
Despite this, \texttt{SHF} provides coresets of comparable quality,
in addition to enabling tractable \iid
sampling, density evaluation, normalization constant bounds, and straightforward
construction via stochastic optimization.

\subsection{Bayesian linear regression}\label{sec:bayesian_lin_reg}
In the setting of Bayesian linear regression, we are given a set of data points
$(x_n,y_n)^N_{n=1}$, each consisting of features $x_n\in\reals^p$ 
and response $y_n\in\reals$, and a model of the form
\[
	\begin{bmatrix} \beta & \log\sigma^2 \end{bmatrix}^T \distas \distNorm(0,I),\quad
	\forall n\in[N], \quad 
y_n \given x_n, \beta, \sigma^2 \distind \distNorm\left(\begin{bmatrix} 1 & x_n^T \end{bmatrix}\beta, \sigma^2\right),
	\label{eq:linreg_model}
\]
where $\beta\in\reals^{p+1}$ is a vector of regression coefficients and
$\sigma^2\in\reals_+$ is the noise variance. The
dataset\footnote{This dataset consists of airport data from \url{https://www.transtats.bts.gov/DL_SelectFields.asp?gnoyr_VQ=FGJ}
and weather data from \url{https://wunderground.com}.} that we use consists of
$N=100,000$ flights, each containing $p=10$ features (e.g., distance of
the flight, weather conditions, departure time, etc), and 
the response variable is the difference, in minutes, between the scheduled and actual
departure times. More details can be found in \cref{supp:linearregdetail}.

\begin{figure*}[t!]
\centering
\begin{subfigure}[b]{0.24\textwidth}
    \scalebox{1}{\includegraphics[width=\textwidth]{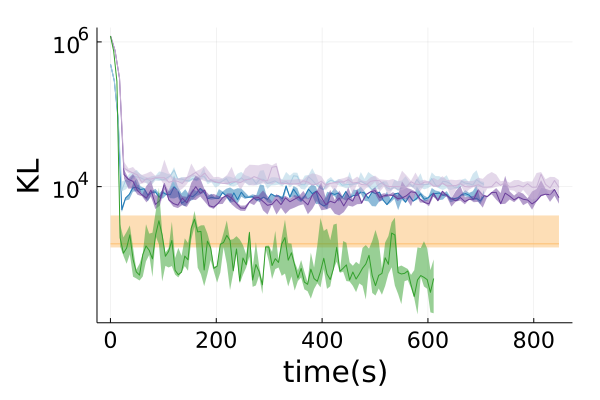}}
    \caption{(a)\label{fig:linearkl}}
\end{subfigure}
\hfill
\centering
\begin{subfigure}[b]{0.24\textwidth}
    \scalebox{1}{\includegraphics[width=\textwidth]{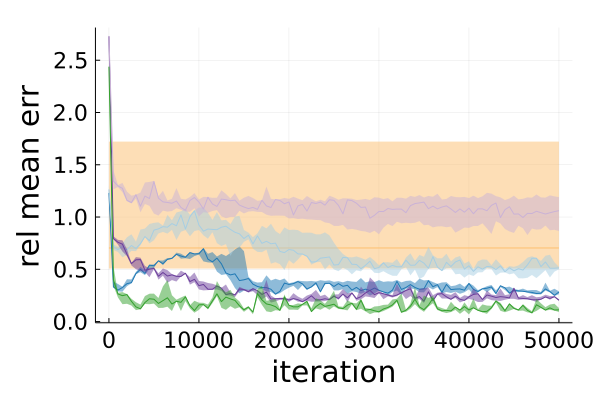}}
    \caption{(b)\label{fig:linearmean}}
\end{subfigure}
\hfill
\centering
\begin{subfigure}[b]{0.24\textwidth}
    \scalebox{1}{\includegraphics[width=\textwidth]{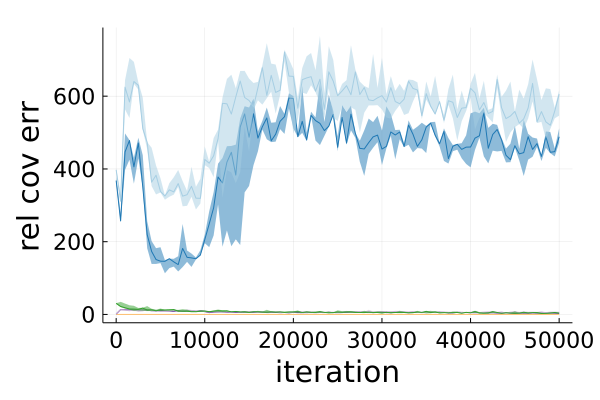}}
    \caption{(c)\label{fig:linearcov}}
\end{subfigure}
\hfill
\centering
\begin{subfigure}[b]{0.24\textwidth}
    \scalebox{1}{\includegraphics[width=\textwidth]{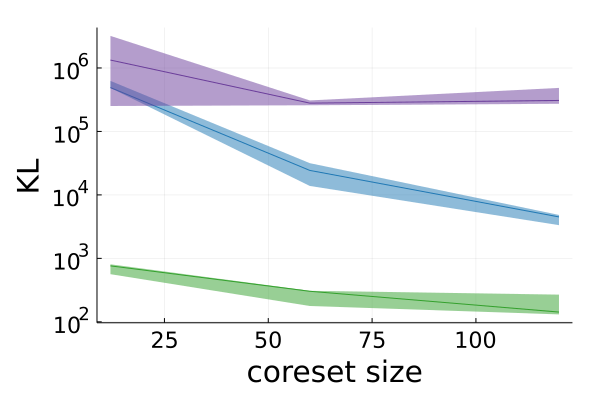}}
    \caption{(d)\label{fig:linearcoresetkl}}
\end{subfigure}
\hfill
\centering
\begin{subfigure}[b]{0.24\textwidth}
    \scalebox{1}{\includegraphics[width=\textwidth]{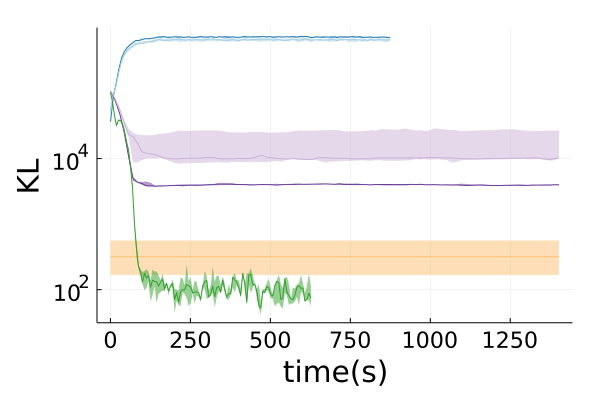}}
    \caption{(e)\label{fig:logiskl}}
\end{subfigure}
\hfill
\centering
\begin{subfigure}[b]{0.24\textwidth}
    \scalebox{1}{\includegraphics[width=\textwidth]{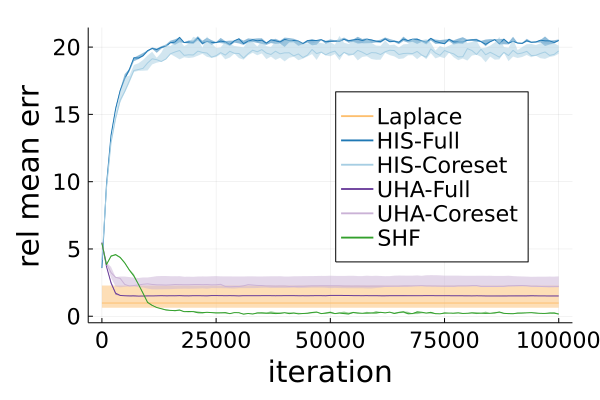}}
    \caption{(f)\label{fig:logismean}}
\end{subfigure}
\hfill
\centering
\begin{subfigure}[b]{0.24\textwidth}
    \scalebox{1}{\includegraphics[width=\textwidth]{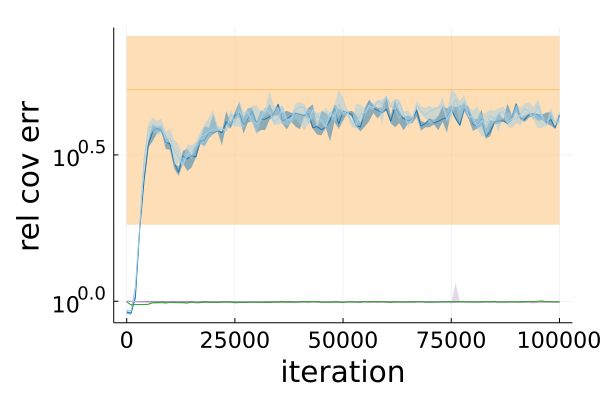}}
    \caption{(g)\label{fig:logiscov}}
\end{subfigure}
\hfill
\centering
\begin{subfigure}[b]{0.24\textwidth}
    \scalebox{1}{\includegraphics[width=\textwidth]{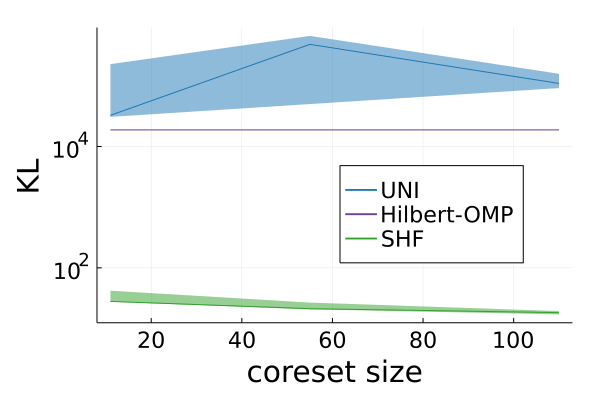}}
    \caption{(h)\label{fig:logiscoresetkl}}
\end{subfigure}

\caption{Linear (top) and logistic (bottom) regression results:
Gaussian approximated KL divergence versus training time (\cref{fig:linearkl,fig:logiskl}),
relative 2-norm mean error (\cref{fig:linearmean,fig:logismean}),
relative Frobenius norm covariance error (\cref{fig:linearcov,fig:logiscov}),
and
Gaussian approximated KL divergence versus coreset size (\cref{fig:linearcoresetkl,fig:logiscoresetkl}). The 
lines indicate the median, and error regions indicate 25$^\text{th}$ to 75$^\text{th}$ 
percentile from 5 runs.}
\label{fig:linearandlogis}
\end{figure*}

Since we no longer have the posterior distribution in closed form, we estimate
the mean and covariance using $5000$ samples from \texttt{Stan}
\cite{carpenter2017stan} and treat them as the true posterior mean and
covariance. \cref{fig:linearkl,fig:linearmean,fig:linearcov} 
show the marginal KL, relative mean error, and relative covariance error of \texttt{SHF}, 
\texttt{HIS}, and \texttt{UHA}, where the marginal KL is estimated using the Gaussian
approximation of the posterior with the estimated mean and covariance. Here we also 
include the posterior approximation obtained using the Laplace approximation as a baseline. 
We see that \texttt{SHF} provides the highest quality posterior approximation.
Furthermore, \cref{fig:linearcoresetkl} shows that \texttt{SHF}
provides a significant improvement in the marginal KL compared with 
competing coreset constructions \texttt{UNI} and \texttt{Hilbert-OMP}.
This is due to the true posterior no longer being
Gaussian; the Laplace approximation 
required by \texttt{Hilbert-OMP} fails to capture the shape of the posterior. 
Additional plots comparing the quality of posterior approximations using various 
other metrics can be found in \cref{supp:linearregdetail}.

\captionsetup[subfigure]{labelformat=empty}

\subsection{Bayesian logistic regression}
In the setting of Bayesian logistic regression, we are given a set of data points 
$(x_n,y_n)^N_{n=1}$, each consisting of features $x_n\in\reals^p$ and label $y_n\in\{0, 1\}$,
and a model of the form
\[
	\forall i\in[p+1],\,\, \beta_i \distiid \distCauchy(0,1),\quad
	\forall n\in[N],\,\,
y_n \distind \distBern\left(\left(1+\exp\left(-\begin{bmatrix} 1 & x_n^T\end{bmatrix} \beta\right)\right)^{-1}\right),
	\label{eq:logreg_model}
\]
where $\beta\in\reals^{p+1}$. The same airline dataset is used with 
the labels indicating whether a flight
is cancelled. Of the flights included, $1.384\%$ were cancelled.
More details can be found in \cref{supp:logisregdetail}.

The same procedures as in the Bayesian linear regression example are followed to generate the results in 
\cref{fig:logiskl,fig:logismean,fig:logiscov,fig:logiscoresetkl}. 
To account for the class imbalance
problem present in the dataset, we construct all subsampled coresets 
with half the data having label $1$ and the
rest with label $0$. The results in
\cref{fig:logiskl,fig:logismean,fig:logiscov,fig:logiscoresetkl} are similar to
those from the Bayesian linear regression example; \texttt{SHF} provides
high quality variational approximations to the posterior.
Additional plots comparing the quality of posterior approximations using various 
other metrics can be found in \cref{supp:logisregdetail}.

\section{Conclusion}\label{sec:conclusion}
This paper introduced sparse Hamiltonian flows, a novel 
coreset-based variational family that enables 
tractable \iid sampling, and evalution of density and normalization constant.
The method randomly subsamples a small set of data, and uses the subsample 
to construct a flow from sparse Hamiltonian dynamics. 
Novel quasi-refreshment steps provide the flow
with the flexibility to match target posteriors 
without introducing additional auxiliary variables. 
Theoretical results show that, in a representative model, the method
can recover the exact posterior using a subsampled dataset 
of the size that is a logarithm of its original size, and that 
quasi-refreshments are guaranteed to reduce the KL divergence to the target.
Experiments demonstrate that the method provides high quality coreset
posterior approximations.
One main limitation of our methodology is that 
the data must be ``compressible" in the sense that 
log-likelihood functions of a subset can be used to well represent the 
full log-likelihood. If the data do not live on some underlying low-dimensional 
manifold, this may not be the case. Additionally, while our 
quasi-refreshment is simple and works well in practice, more work is 
required to develop a wider variety of general-purpose quasi-refreshment 
moves. We leave this for future work.

\begin{ack}
All authors were supported by a National Sciences and Engineering Research Council of Canada
(NSERC) Discovery Grant, NSERC Discovery Launch Supplement, and a gift from Google LLC.
\end{ack}

\small
\bibliographystyle{unsrtnat} %imsart-nameyear
\bibliography{sources}

%%%%%%%%%%%%%%%%%%%%%%%%%%%%%%%%%%%
\section*{Checklist}

\begin{enumerate}

\item For all authors...
\begin{enumerate}
  \item Do the main claims made in the abstract and introduction accurately reflect the paper's contributions and scope?
    \answerYes{}
  \item Did you describe the limitations of your work?
    \answerYes{In \cref{sec:quasi-refresh}, we assume that the momentum distribution is Gaussian. In \cref{sec:conclusion}, we mention that a direction of future work is to develop more general quasi-refreshment moves that do not require such assumptions.}
  \item Did you discuss any potential negative societal impacts of your work?
    \answerNo{This paper presents an algorithm for sampling from a Bayesian posterior distribution in the large-scale data regime. 
     There are possible negative societal impacts
    of downstream applications of this method (e.g.~inference for a particular model and dataset), 
    but we prefer not to speculate about these here.}
  \item Have you read the ethics review guidelines and ensured that your paper conforms to them?
    \answerYes{}
\end{enumerate}

\item If you are including theoretical results...
\begin{enumerate}
  \item Did you state the full set of assumptions of all theoretical results?
    \answerYes{}
        \item Did you include complete proofs of all theoretical results?
    \answerYes{All proofs are provided in the appendix}
\end{enumerate}

\item If you ran experiments...
\begin{enumerate}
  \item Did you include the code, data, and instructions needed to reproduce the main experimental results (either in the supplemental material or as a URL)?
    \answerYes{We have included anonymized code in the supplement.}
  \item Did you specify all the training details (e.g., data splits, hyperparameters, how they were chosen)?
    \answerYes{All details are provided in the main paper and in \cref{supp:expt}.}
        \item Did you report error bars (e.g., with respect to the random seed after running experiments multiple times)?
    \answerYes{All results come with error bars.}
        \item Did you include the total amount of compute and the type of resources used (e.g., type of GPUs, internal cluster, or cloud provider)?
    \answerYes{Hardware is listed at the beginning of the experiments.}
\end{enumerate}

\item If you are using existing assets (e.g., code, data, models) or curating/releasing new assets...
\begin{enumerate}
  \item If your work uses existing assets, did you cite the creators?
    \answerYes{Footnotes to the source URL and bibliography citations are provided.}
  \item Did you mention the license of the assets?
    \answerNo{None of the real datasets we use specify a particular license that the data were released under.}
  \item Did you include any new assets either in the supplemental material or as a URL?
    \answerNA{We do not produce any new assets.}
  \item Did you discuss whether and how consent was obtained from people whose data you're using/curating?
    \answerNo{Our data does not pertain to individuals.}
  \item Did you discuss whether the data you are using/curating contains personally identifiable information or offensive content?
    \answerNo{Our data does not pertain to individuals.}
\end{enumerate}

\item If you used crowdsourcing or conducted research with human subjects...
\begin{enumerate}
  \item Did you include the full text of instructions given to participants and screenshots, if applicable?
    \answerNA{}
  \item Did you describe any potential participant risks, with links to Institutional Review Board (IRB) approvals, if applicable?
    \answerNA{}
  \item Did you include the estimated hourly wage paid to participants and the total amount spent on participant compensation?
    \answerNA{}
\end{enumerate}

\end{enumerate}

%%%%%%%%%%%%%%%%%%%%%%%%%%%%%%%%%%%

\appendix
\section{Additional details on quasi-refreshment}\label{sec:quasi-appendix}
In this section, we extend marginal quasi-refreshment beyond what is 
discussed in the main text and introduce \emph{conditional} quasi-refreshment, 
which tries to match some conditional distribution of $(\rho_t,\theta_t)$ to that 
corresponding distribution in the target.

\paragraph{Marginal quasi-refreshment}
We begin by presenting a more general version of \cref{prop:quasi_marginal}.
\bprop\label{prop:quasi_marginal_general}
Consider random vectors $Y,Z,Y', Z' \in \reals^d$ for some $d\in\nats$. 
Suppose that $Y\indep Z$ and that
we have a bijection $R: \reals^d \to \reals^d$ such that $R(Y')\eqd Y$. Then
\[
\kl{R(Y'),Z'}{Y,Z} &=\kl{Y',Z'}{Y,Z} - \kl{Y'}{Y}.
\]
\eprop
\bprf
Since $R$ is a bijection,
\[
\kl{R(Y'),Z'}{Y,Z} &= \kl{Y',Z'}{R^{-1}(Y),Z}.
\]
Because $Y\indep Z$, and $R(Y')\eqd Y$,
\[
\kl{Y',Z'}{R^{-1}(Y),Z} &= \kl{Y'}{R^{-1}(Y)} + \EE\left[\kl{(Z'|Y')}{Z}\right]\\
&= \EE\left[\kl{(Z'|Y')}{Z}\right].
\]
Then we add and subtract $\kl{Y'}{Y}$ to obtain the final result,
\[
\EE\left[\kl{(Z'|Y')}{Z}\right] &= \kl{Y',Z'}{Y,Z} - \kl{Y'}{Y}.
\]
\eprf
Then \cref{prop:quasi_marginal} follows immediately from 
\cref{prop:quasi_marginal_general}.
\bprfof{\cref{prop:quasi_marginal}}
By setting $Y = \rho, Z = \theta, Y'=\rho_t$, and $Z' = \theta_t$, we 
arrive at the stated result.
\eprfof
More generally, in order to apply \cref{prop:quasi_marginal_general},
we first need to split the momentum variable into 
two components $(\rho^{(1)}, \rho^{(2)})$ in such a way that
$\rho^{(1)}\indep \rho^{(2)}$ under $(\theta,\rho)\distas\bpi$, and then
set $Y = \rho^{(1)}$ and $Z = (\rho^{(2)}, \theta)$.
Since we know that $\rho\distas \distNorm(0, I)$, we have quite a few options.
For example:
\bitems
\item Set $Y = \rho$, and $Z = \theta$. Then
\[
Y\indep Z \quad\text{and}\quad Y\distas\distNorm(0, I).
\]
\item Set $Y = \|D\rho\|^2_2$ and $Z =\left( \frac{D\rho}{\|D\rho\|_2}, (I-D)\rho, \theta\right)$
for any binary diagonal matrix $D\in\reals^{d\times d}$. Then 
\[
Y\indep Z \quad\text{and}\quad Y\distas\distChiSq(\tr D).
\]
\item Set $Y = a^T\rho$ for any $a\in\reals^d$ such that $\|a\|=1$,
and $Z = \left((I-aa^T)\rho, \theta\right)$.
Then
\[
Y\indep Z\quad\text{and}\quad Y\distas\distNorm(0,1).
\]
\eitems
Note again that the first option recovers the marginal quasi-refreshment in the main text.
Now we use the above decomposition to design a marginal quasi-refreshment move.
Suppose we have run the weighted sparse leapfrog integrator \cref{eq:wleapfrog} up to time $t$,
resulting in a current random state $\theta_t, \rho_t$.
Let the decomposition of the current 
state $(\theta_t,\rho_t)$ that we select above be denoted $Y', Z'$.
Then the marginal quasi-refreshment move involves finding a map
$R$ such that $R(Y') \eqd Y$. We can then refresh the state via
$(R(Y'), Z')$, and continue the flow.
In addition to the parametric approach to designing $R$ presented  in the main text,
if $Y$ is 1-dimensional---for example, if we are trying to refresh the
momentum norm $Y=\|\rho\|_2$---then given that we know the CDF $F$ of $Y$,
we can estimate the CDF $\hF$ of $Y'$ using samples from the flow
at timestep $t$, and then set
\[
R(x) = F^{-1}\left(\hF(x)\right).
\]
For example, we could use this inverse CDF map technique
to refresh the distribution of $\|\rho\|^2_2$ back to $\distChiSq(d)$,
or to refresh the distribution of $a^T\rho$ back to $\distNorm(0,1)$.

\paragraph{Conditional quasi-refreshment}
A conditional quasi-refreshment move is one
that tries to make some conditional distribution of $(\rho_t,\theta_t)$
match the same conditional in the target. If one is able to accomplish this
for \emph{any} conditional distribution (with no requirement of independence
as in the marginal case), the KL divergence is guaranteed to reduce.
\bprop\label{prop:quasi_conditional}
Consider random vectors $Y,Z,Y',Z' \in \reals^d$ for some $d\in\nats$.
Suppose for each $s\in\reals^d$ we have a bijection
$R_s : \reals^d\to\reals^d$ such that
$(R_s(Y') \given Z'=s) \eqd (Y \given Z=s)$. Then
\[
\kl{R_{Z'}(Y'), Z'}{Y,Z} &= \kl{Z'}{Z}.
\]
\eprop
\bprf
By assumption the distribution of $Y$ given $Z=s$
is the same as that of $R_s(Y')$ given $Z'=s$; so
the result follows directly from the decomposition
\[
\kl{R_{Z'}(Y'), Z'}{Y, Z} &= \kl{Z'}{Z} + \EE\left[\kl{(R_S(Y') | Z'=S)}{(Y | Z=S)}\right], \quad S\eqd Z'\\
&= \kl{Z'}{Z}.
\]
\eprf
Conditional moves are much harder to
design than marginal moves in general. One case of particular utility occurs when one is willing
to assume that $(\theta_t, \rho_t)$ are roughly jointly normally distributed.
In this case,
\[
\begin{bmatrix}\theta_t \\ \rho_t\end{bmatrix} \distas\distNorm\left(\begin{bmatrix} \mu_\theta \\ \mu_\rho\end{bmatrix}, 
\begin{bmatrix} \Sigma_{\theta\theta} & \Sigma_{\theta\rho}\\
      \Sigma_{\theta\rho}^T & \Sigma_{\rho\rho}\end{bmatrix}\right),
\]
so one can refresh the momentum $\rho_t$ by updating it to $R_{\theta_t}(\rho_t)$, where
\[
R_s(x) = \Sigma^{-1/2}\left(x - \mu_\rho - \Sigma_{\theta\rho}^T\Sigma_{\theta\theta}^{-1}\left(s - \mu_\theta\right)\right) \quad\text{and}\quad
\Sigma = \Sigma_{\rho\rho} - \Sigma_{\theta\rho}^T\Sigma_{\theta\theta}^{-1}\Sigma_{\theta\rho}.
\]
Here \cref{prop:quasi_conditional} applies by setting $Y=\rho, Z=\theta, Y'=\rho_t$, and $Z'=\theta_t$.
In order to use this quasi-refreshment move, one can either include the covariance matrices and 
mean vectors as tunable parameters in the optimization, or use samples from the flow
at step $t$ to estimate them directly.

\section{Details of experiments} \label{supp:expt}

We begin by describing in detail the differences between \texttt{HIS-Full}/\texttt{UHA-Full} and
\texttt{HIS-Coreset}/\texttt{UHA-Coreset}.
It is worth noting noting that we were unable to train \texttt{HIS}/\texttt{UHA} with full-dataset
flow dynamics; even on a small 2-dimensional Gaussian location model with 
100 data points, these methods took over 8 minutes for training.
Therefore, as suggested by
\citep{Caterini18,Geffner21}, we train the leapfrog step sizes and annealing
parameters by using a random minibatch of the data in each iteration to
construct the flow dynamics. To then obtain valid ELBO estimates for comparison,
we generate samples from the trained flow with leapfrog transformations
based on the full dataset (\texttt{HIS-Full}/\texttt{UHA-Full}).
As a simple heuristic baseline that also provides a valid ELBO estimate, we 
compare to the trained flow with leapfrog transformations
based on a fixed, uniformly sampled coreset (\texttt{HIS-Coreset}/\texttt{UHA-Coreset}) 
of the same size as used for \texttt{SHF}. 

We now describe the detailed settings that apply across all three experiments
in the main text.
In all experiments, \texttt{SHF} uses the 
quasi-refreshment from \cref{eq:marginal_standardization}
initialized using the warm-start procedure in \cref{sec:sparse_flows} with a batch of $100$ samples. 
We use the same number of
leapfrog, tempering, and (quasi-)refreshment steps for all of
\texttt{SHF}, \texttt{HIS} and \texttt{UHA} respectively.
The leapfrog step sizes are initialized at the same value
for all three methods. 
For all three methods (\texttt{SHF}, \texttt{HIS}, and \texttt{UHA}), the unnormalized log target density
used in computing the ELBO objective is estimated using a minibatch
of $S=100$ data points for each optimization iteration.
For both \texttt{HIS} and \texttt{UHA}, the leapfrog transitions themselves are also based on a
fresh uniformly sampled minibatch of size 30---the same as the coreset size for \texttt{SHF}---at 
each optimization iteration. \texttt{HIS} and \texttt{UHA} both also involve
tempering procedures, requiring optimization over the tempering schedule $0 \leq
\beta_1 \leq \dots \leq \beta_{R - 1} \leq \beta_R = 1$, where $R$ denotes the
number of tempering steps (equal to the number of quasi-refreshment / refreshment steps in all methods). 
We consider a
reparameterization of $(\beta_1, \dots, \beta_R)$ to $(\alpha_1, \dots,
\alpha_{R-1})$, where $\alpha_r = \text{Logit}(\beta_{r + 1}/\beta_r)$ 
for $r \in \{1, \dots, R-1\}$, ensuring a set of unconstrained parameters. The initial value
for each $r$ is set to $\alpha_r = 1$.  
For \texttt{UHA}, the initial damping coefficient of its partial
momentum refreshment is set to $0.5$. 
We estimate all evaluation metrics using $100$ samples.
To estimate KSD, we use the IMQ base kernel with its parameters set to the same values as outlined in 
\cite{gorham2017measuring} ($\beta = -\frac{1}{2}$ and $c=1$).

In addition to \texttt{HIS} and \texttt{UHA}, we also include adaptive HMC and NUTS in our 
comparison of density evaluation and sample generation times.
We tune adaptive HMC and NUTS with a target acceptance rate of $0.65$
during a number of burn-in iterations equal to the number of output samples,
and include the burn-in time in the timing results for these two methods.
Finally, we compare the quality of coresets constructed by \texttt{SHF} against \texttt{UNI}
and \texttt{Hilbert-OMP}. For both \texttt{UNI} and \texttt{Hilbert-OMP}, we use NUTS to 
draw samples from the coreset posterior approximation.
For \texttt{Hilbert-OMP}, we use a random log-likelihood function projection of
dimension $100d$, where the true posterior parameter is of dimension $d$,
generated from the Laplace approximation \cite{tierney1986accurate}.

\subsection{Synthetic Gaussian} \label{supp:syndetail}

To train \texttt{SHF}, a total of 5 quasi-refreshments are used
with 10 leapfrog steps in between; a similar schedule is used in
\texttt{HIS} and \texttt{UHA} for momentum tempering and refreshment
respectively. 
The initial distribution is set to $\theta_0 \distas \distNorm(0, I)$ and $\rho_0 \distas \distNorm(0, I)$.
For all methods, the number of optimization
iterations is set to $20,000$, and the initial leapfrog step size is set to $0.01$ across all $d$ dimensions. 
We train all methods with ADAM with initial learning rate $0.001$.
\subsection{Bayesian linear regression} \label{supp:linearregdetail}

To train \texttt{SHF}, a total of 8 quasi-refreshments are used
with 10 leapfrog steps in between; a similar schedule is used in
\texttt{HIS} and \texttt{UHA} for momentum tempering and refreshment
respectively. 
The initial distribution is set to $\theta_0 \distas \distNorm(15, 0.01I)$ and $\rho_0 \distas \distNorm(0, I)$.
For all methods, the number of optimization
iterations is set to $50,000$, and the initial leapfrog 
step size is set to $0.02$ across all $d$ dimensions except for the dimension for the $\log\sigma^2$ term, where
the step size is set to $0.0002$. We train all methods with ADAM with initial learning rate $0.002$.
We also include the posterior approximation obtained from the Laplace approximation, where we search for 
the mode of the target density at some location generated from the same distribution as that of $\theta_0$.
\cref{fig:linearlogisadditional,fig:linearlogistime,fig:linearlogiscoreset} provide additional results for this experiment.
\cref{fig:linearelbo} shows that the ELBO obtained from \texttt{SHF} is a
tighter lower bound of the log normalization constant compared to \texttt{HIS}
and \texttt{UHA}. \cref{fig:lineared} shows that \texttt{SHF} produces
a posterior approximation in terms of energy distance than all other methods; \cref{fig:linearksd} shows that 
\texttt{SHF} is competitive, if not better, than the other methods in terms of KSD.
\cref{fig:linearcoresetmean,fig:linearcoresetcov,fig:linearcoreseted,fig:linearcoresetksd} show that the coreset 
constructed by \texttt{SHF} is of better quality than those obtained from \texttt{UNI} and \texttt{Hilbert-OMP} 
in terms of all four metrics shown.
Finally, \cref{fig:lineardensity,fig:linearsample} show the computational gain of using
\texttt{SHF} to approximate density and generate posterior samples as compared to
\texttt{HIS} and \texttt{UHA}.

\subsection{Bayesian logistic regression} \label{supp:logisregdetail}

\begin{figure*}[t!]
    \centering 
\begin{subfigure}[b]{.32\textwidth} 
    \scalebox{1}{\includegraphics[width=\textwidth]{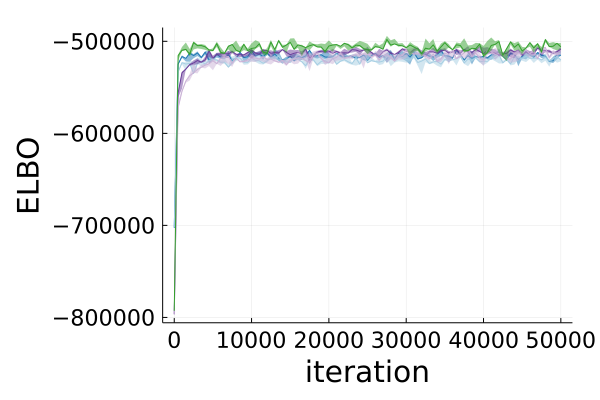}}
    \caption{(a)\label{fig:linearelbo}}
\end{subfigure}
\hfill
\centering 
\begin{subfigure}[b]{.32\textwidth} 
    \scalebox{1}{\includegraphics[width=\textwidth]{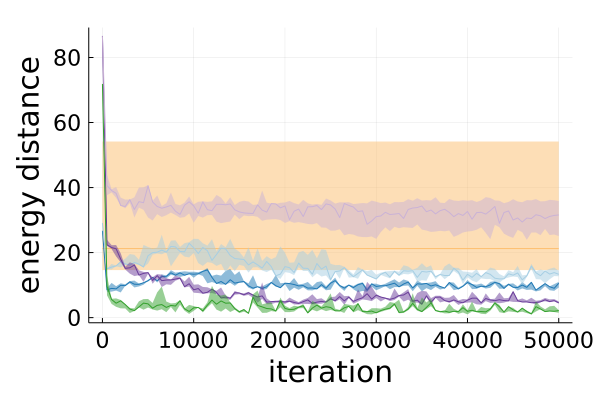}}
    \caption{(b)\label{fig:lineared}}
\end{subfigure}
\hfill
\centering 
\begin{subfigure}[b]{.32\textwidth} 
    \scalebox{1}{\includegraphics[width=\textwidth]{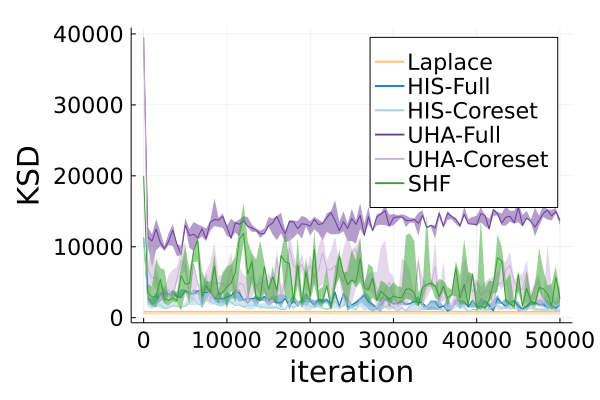}}
    \caption{(c)\label{fig:linearksd}}
\end{subfigure}
\hfill
\centering 
\begin{subfigure}[b]{.32\textwidth} 
    \scalebox{1}{\includegraphics[width=\textwidth]{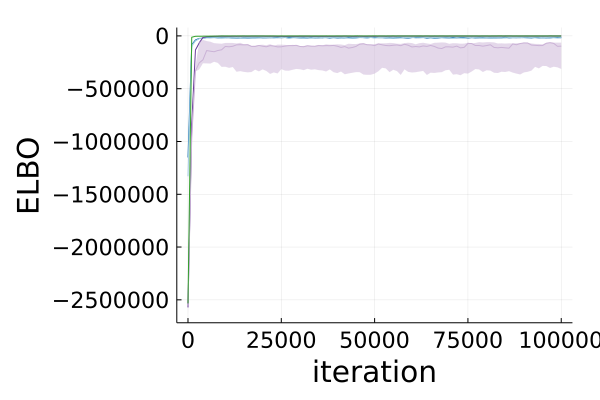}}
    \caption{(d)\label{fig:logisticelbo}}
\end{subfigure}
\hfill
\centering 
\begin{subfigure}[b]{.32\textwidth} 
    \scalebox{1}{\includegraphics[width=\textwidth]{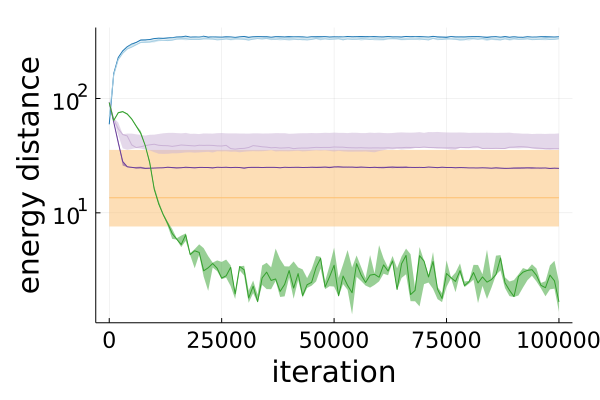}}
    \caption{(e)\label{fig:logisticed}}
\end{subfigure}
\hfill
\centering 
\begin{subfigure}[b]{.32\textwidth} 
    \scalebox{1}{\includegraphics[width=\textwidth]{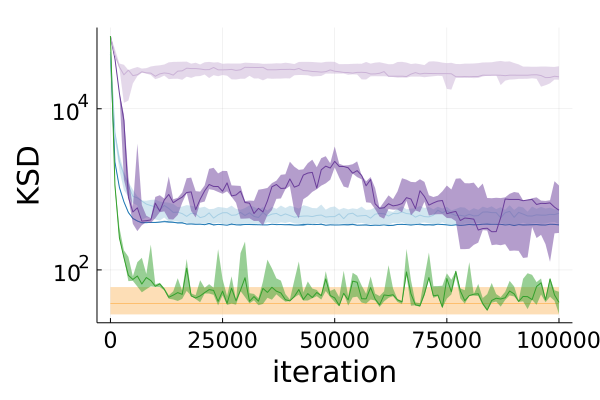}}
    \caption{(f)\label{fig:logisticksd}}
\end{subfigure}
\hfill

\caption{Linear (top row, \cref{fig:linearelbo,fig:lineared,fig:linearksd}) and 
logistic (bottom row \cref{fig:logisticelbo,fig:logisticed,fig:logisticksd}) regression: posterior approximation quality results. The lines indicate the median, and error regions indicate 25$^\text{th}$ to 75$^\text{th}$ percentile from 5 runs.}
\label{fig:linearlogisadditional}
\end{figure*}

\begin{figure*}[t!]
\centering 
\begin{subfigure}[b]{.24\textwidth} 
    \scalebox{1}{\includegraphics[width=\textwidth]{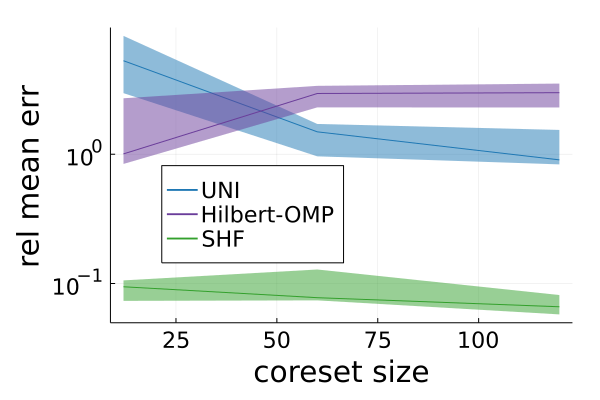}}
    \caption{(a)\label{fig:linearcoresetmean}}
\end{subfigure}
\hfill
\centering 
\begin{subfigure}[b]{.24\textwidth} 
    \scalebox{1}{\includegraphics[width=\textwidth]{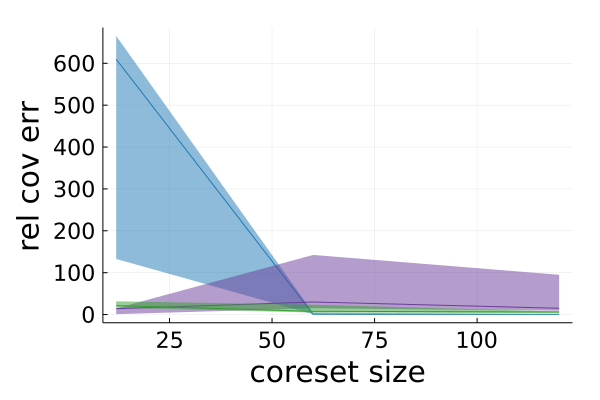}}
    \caption{(a)\label{fig:linearcoresetcov}}
\end{subfigure}
\hfill
\centering 
\begin{subfigure}[b]{.24\textwidth} 
    \scalebox{1}{\includegraphics[width=\textwidth]{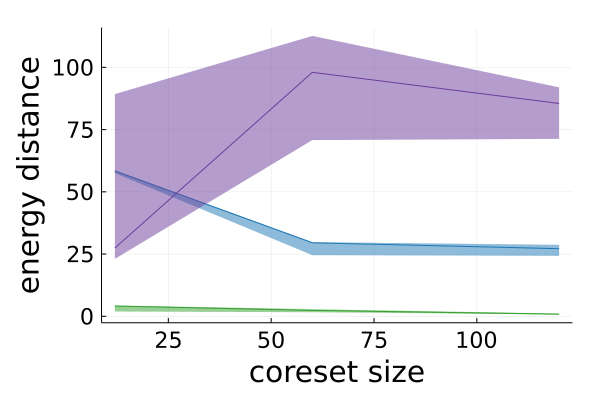}}
    \caption{(a)\label{fig:linearcoreseted}}
\end{subfigure}
\hfill
\centering 
\begin{subfigure}[b]{.24\textwidth} 
    \scalebox{1}{\includegraphics[width=\textwidth]{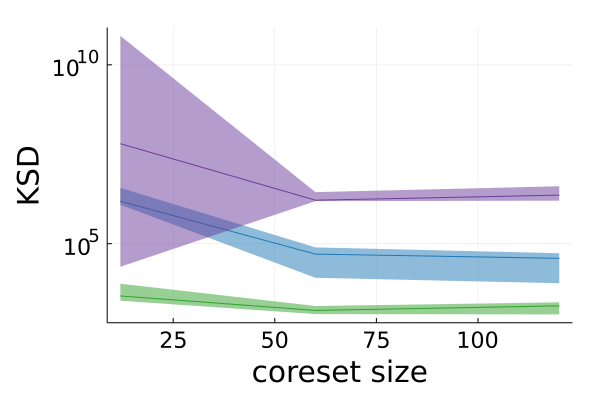}}
    \caption{(a)\label{fig:linearcoresetksd}}
\end{subfigure}
\hfill
\centering 
\begin{subfigure}[b]{.24\textwidth} 
    \scalebox{1}{\includegraphics[width=\textwidth]{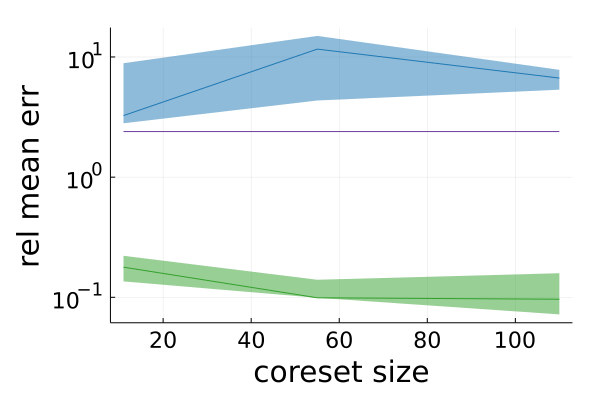}}
    \caption{(a)\label{fig:logisticcoresetmean}}
\end{subfigure}
\hfill
\centering 
\begin{subfigure}[b]{.24\textwidth} 
    \scalebox{1}{\includegraphics[width=\textwidth]{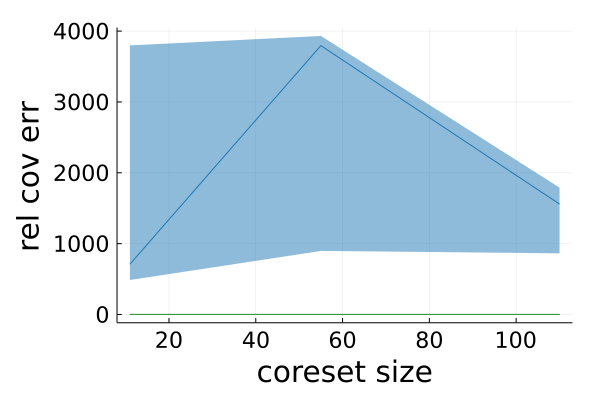}}
    \caption{(a)\label{fig:logisticcoresetcov}}
\end{subfigure}
\hfill
\centering 
\begin{subfigure}[b]{.24\textwidth} 
    \scalebox{1}{\includegraphics[width=\textwidth]{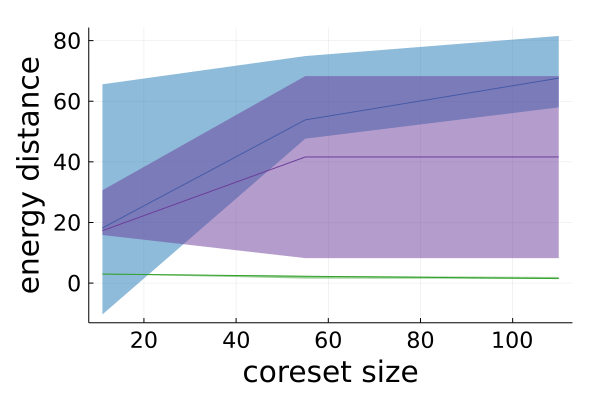}}
    \caption{(a)\label{fig:logisticcoreseted}}
\end{subfigure}
\hfill
\centering 
\begin{subfigure}[b]{.24\textwidth} 
    \scalebox{1}{\includegraphics[width=\textwidth]{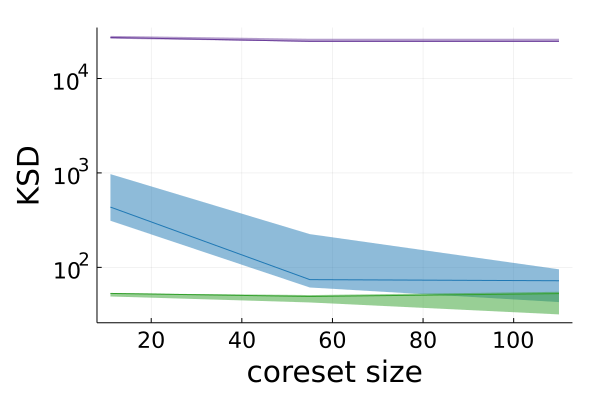}}
    \caption{(a)\label{fig:logisticcoresetksd}}
\end{subfigure}
\hfill

\caption{Linear (top row, \cref{fig:linearcoresetmean,fig:linearcoresetcov,fig:linearcoreseted,fig:linearcoresetksd}) and 
logistic (bottom row \cref{fig:logisticcoresetmean,fig:logisticcoresetcov,fig:logisticcoreseted,fig:logisticcoresetksd}) regression: posterior approximation quality results. The lines indicate the median, and error regions indicate 25$^\text{th}$ to 75$^\text{th}$ percentile from 5 runs.}
\label{fig:linearlogiscoreset}
\end{figure*}

\begin{figure*}[t!]
\centering 
\begin{subfigure}[b]{.24\textwidth} 
    \scalebox{1}{\includegraphics[width=\textwidth]{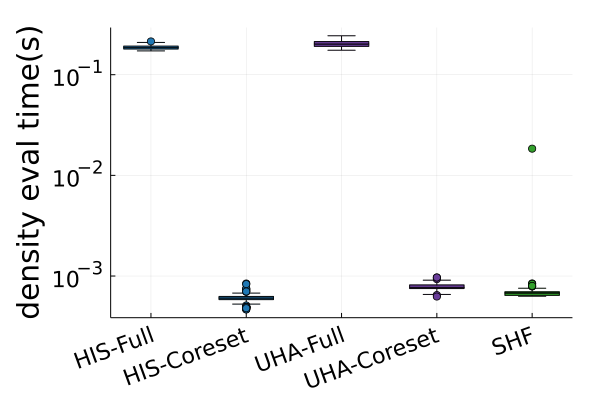}}
    \caption{(a)\label{fig:lineardensity}}
\end{subfigure}
\hfill
\centering
\begin{subfigure}[b]{0.24\textwidth}
    \scalebox{1}{\includegraphics[width=\textwidth]{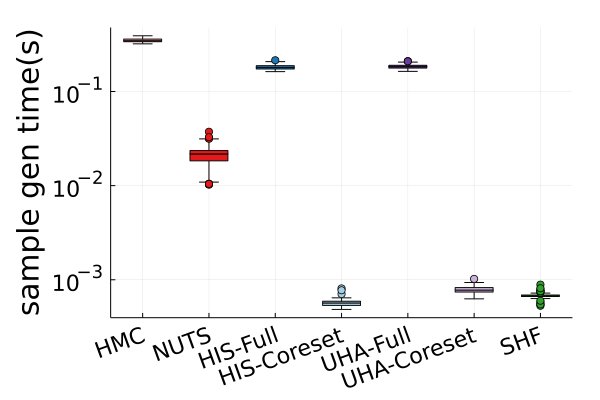}}
    \caption{(b)\label{fig:linearsample}}
\end{subfigure}
\hfill    
\centering 
\begin{subfigure}[b]{.24\textwidth} 
    \scalebox{1}{\includegraphics[width=\textwidth]{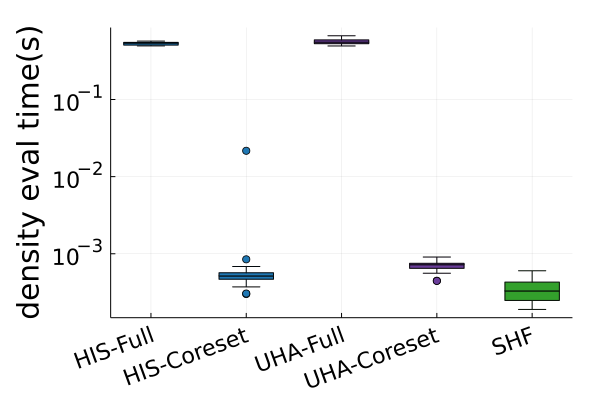}}
    \caption{(c)\label{fig:logisdensity}}
\end{subfigure}
\hfill
\centering
\begin{subfigure}[b]{0.24\textwidth}
    \scalebox{1}{\includegraphics[width=\textwidth]{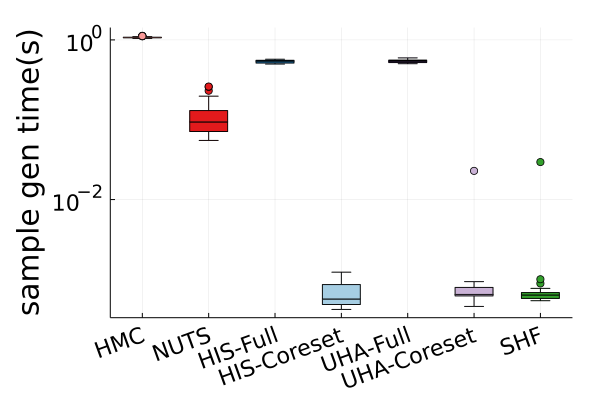}}
    \caption{(d)\label{fig:logissample}}
\end{subfigure}

\caption{Linear (\cref{fig:lineardensity,fig:linearsample}) and logistic (\cref{fig:logisdensity,fig:logissample}) regression: timing results based on 100 samples.}
\label{fig:linearlogistime}
\end{figure*}

To train \texttt{SHF}, a total of 8 quasi-refreshments are used
with 10 leapfrog steps in between; a similar schedule is used in
\texttt{HIS} and \texttt{UHA} for momentum tempering and refreshment
respectively. 
The initial distribution is set to $\theta_0 \distas \distNorm(15, 10^{-4}I)$ and $\rho_0 \distas \distNorm(0, I)$.
For all methods, the number of optimization
iterations is set to $100,000$, and the initial leapfrog 
step size is set to $0.0005$ across all $d$ dimensions. 
We train all methods with ADAM with initial learning rate $0.001$.
We also include the posterior approximation obtained from the Laplace approximation, where we search for 
the mode of the target density at some location generated from the same distribution as that of $\theta_0$.
\cref{fig:linearlogisadditional,fig:linearlogiscoreset,fig:linearlogistime} provide additional results for this experiment, 
where similar conclusions as in the Bayesian linear regression experiment can be drawn.

\captionsetup[subfigure]{labelformat=empty}

\captionsetup[subfigure]{labelformat=empty}

\captionsetup[subfigure]{labelformat=empty}
\section{Gaussian KL upper bound proof} \label{sec:upperbound_proof}
\bprfof{\cref{prop:subsample_basis}}
Suppose we are given a particular choice $\mcI \subseteq [N]$ of $N$ indices
of size $|\mcI| = M \in \nats$, $M \leq N$.
Let $\mcW_\mcI = \left\{w \in\reals_+^N : \forall n \in[N], n\notin\mcI \implies w_n = 0\right\}$.
In the $d$-dimensional normal location model, the exact and $w\in\mcW_\mcI$-coreset posteriors
are multivariate Gaussian distributions, denoted as $\distNorm{\left(\mu_1, \Sigma_1\right)}$ and 
$\distNorm{\left(\mu_{w}, \Sigma_{w}\right)}$ respectively, with mean and covariance 
\[
\Sigma_{1}=\frac{1}{1+ N} I, \quad \mu_{1}=\Sigma_{1}\left( \sum_{n=1}^{N} X_{n}\right)
\quad \text{and} \quad
\Sigma_{w}\!=\!\frac{I}{1+ \left(\sum_{n\in\mcI} w_n\right)}, 
\quad
\mu_{w}\!=\!\Sigma_{w}\left( \sum_{n\in\mcI} w_n X_{n}\right). \label{eq:gauss_posts}
\]
The KL divergence between these two distributions is
\[ 
\kl{\pi_{w}}{\pi}
=& \frac{1}{2} \left[ -d\log \left( \frac{1+ N}{1+ \sum_{n\in\mcI} w_n}\right) - d  + d \left( \frac{1+ N}{1+ \sum_{n\in\mcI} w_n}\right)
+  (\mu_{1} - \mu_{w})^T \Sigma_{1}^{-1}(\mu_{1} - \mu_{w})\right]. \label{eq:KL_from_coreset_post_to_exact}
\]
We can bound this quantity above by adding the constraint $\sum_{n\in\mcI} w_n = N$, yielding
\[ 
\min_{w\in\mcW_\mcI} \kl{\pi_w}{\pi} \leq
\min_{w\in\triangle^{M-1}} 
\frac{N^2}{2(N+1)} \left\|\bX - \sum_{n\in\mcI}w_n X_n\right\|^2,
\]
where $\bX = \frac{1}{N}\sum_{n=1}^N X_n$.
and $\triangle^{M-1}$ is the $M-1$-dimensional simplex $w\geq 0$, $1^Tw = 1$.
We aim to show that with high probability (over uniform random choice of $\mcI$ and realizations of $X_n, n\in [N]$),
there exists a $w\in\triangle^{M-1}$ such that $\bX = \sum_{n\in\mcI}w_nX_n$, and hence the optimal KL divergence is 0.

Since $X_n \distiid \distNorm(0, I)$, $\sqrt{N}\bX \distas\distNorm(0, I)$, so $\|\sqrt{N}\bX\|^2 \distas \distChiSq(d)$ and therefore
for any $s > \sqrt{d/N}$,
\[
\Pr\left(\|\bX\| > s \right) &\leq \left(\frac{s^2N}{d}e^{1-\frac{s^2N}{d}}\right)^{d/2}. \label{eq:bxbound}
\]
In other words, as $N$ increases, we can expect $\bX$ to concentrate around the origin.
Therefore as long as the convex hull of $X_n$, $n\in\mcI$ 
contains a ball of some fixed radius around the origin with high probability, we know that 
$\bX$ is a convex combination of $X_n, n\in\mcI$.
The radius of the largest origin-centered ball inside the convex hull of $X_n, n\in\mcI$ can be expressed as
\[
r^\star &= \min_{a \in \reals^d : \|a\|=1, b \geq 0} \,  b\quad \text{s.t.}\quad  \forall n\in\mcI, \quad a^TX_n - b \leq 0 
= \min_{a \in \reals^d : \|a\|=1} \, \max_{n\in\mcI}\, a^TX_n.
\]

By \citet[Corollary 1.2]{Boroczky03}, $S^d$ can be covered by
\[
N_d(\phi) = \frac{C\cdot \cos\phi}{\sin^d\phi} d^{\frac{3}{2}}\log(1+d\cos^2\phi) \leq \phi^{-d}A_d, \quad A_d = C e^{\frac{d}{2}}d^{\frac{3}{2}}\log(1+d).\label{eq:Nphid}
\]
balls of radius $0 < \phi \leq \arccos\frac{1}{\sqrt{d+1}}$, where $C$ is a universal constant.
Denote the centres of these balls $a_i \in S^d$, $i=1,\dots, N_d(\phi)$. Then
\[
r^\star &\geq \min_{i \in [N_d(\phi)], v \in \reals^d : \|v\|\leq \phi} \, \max_{n\in\mcI}\, (a_i+v)^TX_n\\
&\geq \min_{i \in [N_d(\phi)]} \, \max_{n\in\mcI}\, a_i^TX_n - \phi\|X_n\|.
\]
Therefore the probability that the largest origin-centred ball enclosed in the convex
hull is small is bounded above by
\[
\Pr\left(r^\star \leq t\right) &\leq
\Pr\left(\min_{i \in [N_d(\phi)]} \, \max_{n\in\mcI}\, a_i^TX_n - \phi\|X_n\| \leq t\right)\\
&\leq N_d(\phi)\Pr\left(\max_{n\in\mcI}\, a_i^TX_n - \phi\|X_n\| \leq t\right)\\
&= N_d(\phi)\Pr\left(a^TZ - \phi\|Z\| \leq t\right)^M,
\]
for $a \in S^d$ and $Z \distas \distNorm(0, I)$.
Since $Z$ has a spherically symmetric distribution, $a$ is arbitrary, so we can choose $a = \bmat 1 & 0 & \dots & 0 \emat^T$.
If we let $U \distas\distNorm(0,1)$ and $V\distas\distChiSq(d-1)$ be independent, this yields
\[
\Pr\left(a^TZ - \phi\|Z\| \leq t\right) &= \Pr\left(U - \phi\sqrt{U^2+V} \leq t\right)\\
&= \Pr\left(U-t \leq \phi\sqrt{U^2+V}\right)\\
&\leq \Pr\left(U < 2t\right) + \Pr\left(V \geq \phi^{-2}(U-t)^2 - U^2, U \geq 2t\right)\\
&\leq \Phi(2t) + \Pr\left(V' \geq \phi^{-2}t^2\right),
\]
where $V'\distas\distChiSq(d)$ and $\Phi(\cdot)$ is the CDF of the standard normal.
Therefore as long as $t > \phi\sqrt{d}$,
\[
\Pr\left(V' \geq \phi^{-2}t^2\right) &=\Pr\left(V' \geq d\left(\frac{t}{\phi\sqrt{d}}\right)^2\right)\\
&\leq \left(\left(\frac{t}{\phi\sqrt{d}}\right)^2e^{1-\left(\frac{t}{\phi\sqrt{d}}\right)^2}\right)^{d/2}.
\]
We now combine the above results to show that for any $t > \phi\sqrt{d}$,
\[
\Pr\left(r^\star \leq t\right)&\leq N_d(\phi)\left(\Phi(2t) + d^{-\frac{d}{2}}e^{\frac{d}{2}}\phi^{-d}t^de^{-\frac{1}{2}\phi^{-2}t^2}\right)^M. \label{eq:rbound}
\]
Finally we combine  the bound on the norm of $\bX$ \cref{eq:bxbound}, the bound on $r^\star$ \cref{eq:rbound}, 
and the covering number of $S^d$ \cref{eq:Nphid} for the final result.
For any $t > \max\{\phi\sqrt{d}, \sqrt{d/N}\}$ and $\phi \leq \arccos\frac{1}{\sqrt{d+1}}$,
\[
\Pr\left(\bX \notin\conv(X_n)_{n\in\mcI}\right)
&\leq d^{-\frac{d}{2}}e^{\frac{d}{2}}t^d N^{d/2}e^{-\frac{1}{2}t^2N} +
 \phi^{-d}A_d\left(\Phi(2t) + d^{-\frac{d}{2}}e^{\frac{d}{2}}\phi^{-d}t^de^{-\frac{1}{2}\phi^{-2}t^2}\right)^M
\]
Set $\phi = \sqrt{\frac{1}{N}}$ and $t = s\sqrt{d/N}$.
As long as $N \geq 2$ and $s > 1$, we are guaranteed that
$t >\max\{\phi\sqrt{d}, \sqrt{d/N}\}$ and $\phi \leq \arccos\frac{1}{\sqrt{d+1}}$ as required, and so
\[
\Pr\left(\bX \notin\conv(X_n)_{n\in\mcI}\right)
&\leq e^{\frac{d}{2}}s^d e^{-\frac{d}{2}s^2} +
N^{\frac{d}{2}}A_d\left(\Phi\left(s\sqrt{\frac{4d}{N}}\right) + e^{\frac{d}{2}}s^de^{-\frac{d}{2}s^2}\right)^M.
\]
If we set $s = \sqrt{\log N + 1}$, 
\[
e^{\frac{d}{2}}s^d e^{-\frac{d}{2}s^2} = N^{-\frac{d}{2}}(\log N+1)^{\frac{d}{2}} \quad \Phi\left(s\sqrt{\frac{4d}{N}}\right) = \frac{1}{2} + O\left(\sqrt{\frac{\log N}{N}}\right).
\]
So then setting $M = d\log_2(N) - \frac{d}{2}\log_2(\log(N)) + \log_2 A_d$ yields the claimed result.
\eprfof

\section{Gaussian KL lower bound proof}\label{sec:lowerbound_proof}

\bprfof{\cref{prop:problem}}
Let $z(t) = (\theta_t, \rho_t)$. 
The evolution of $z(t)$ is determined by a time-inhomogeneous linear
ordinary differential equation,
\[
\der{z(t)}{t} &= A(t) z(t)  \qquad A(t) = \begin{bmatrix} 0 & 1 \\ -\sigma^{-2} & -\gamma(t)\end{bmatrix},
\]
with solution
\[
z(t) &= e^{B(t)} z(0) \qquad B(t) = \begin{bmatrix} 0 & t \\ -\sigma^{-2}t & g(t)\end{bmatrix}, \qquad 
g(t) = -\int_0^t\gamma(t)\dee t. 
\]
Therefore by writing $\bpi_0 = \distNorm\left( m(0), \Sigma(0) \right)$, $z(t) \distas q_t = \distNorm\left( m(t), \Sigma(t) \right)$, where
\[
m(t) &= e^{B(t)}m(0) \qquad \Sigma(t) = e^{B(t)}\Sigma(0)e^{B(t)^T}. \label{eq:mtsigt}
\]
The second result follows because tempered Hamiltonian dynamics where $\gamma(t) = 0$ identically is just standard Hamiltonian dynamics.
For the first result, suppose $q_0 = \distNorm \left(\begin{bmatrix} \mu \\
0 \end{bmatrix}, \begin{bmatrix} 1 & 0 \\ 0 & \beta^2\end{bmatrix}\right)$ for some
$\mu\in\reals, \beta\in\reals_+$. Note that in this case, it suffices to
consider 
\[
m(0) = \begin{bmatrix} \mu\\ 0\end{bmatrix} \qquad \Sigma(0) = I.\label{eq:init_conds}
\] 
This is because the function $\gamma(t)$ is arbitrary; one can, for example, set $\gamma(t) = -\epsilon^{-1}\log \beta$ for $t\in[0, \epsilon)$
for an arbitrarily small $\epsilon > 0$, such that the state at time $\epsilon$ is arbitrarily close to the desired initial condition. The KL divergence from $q_t$ to $\bpi$ is
\[
\kl{q_t}{\bpi} &= \frac{1}{2}\left[ \log\left( \frac{\det \Sigma}{\det\Sigma(t)} \right) - 2 + \tr\left(\Sigma^{-1}\Sigma(t)\right) + m(t)^T \Sigma^{-1}m(t) \right]\\
&=\frac{1}{2}\left[ \log\left( \frac{\det \Sigma}{\det\Sigma(t)} \right) - 2 + \tr\left(\Sigma^{-1}\left(\Sigma(t)+m(t)m(t)^T\right)\right) \right]. \label{eq:KL_initial}
\]
By \cref{eq:mtsigt,eq:init_conds} and the identity $\tr A^TA = \|A\|_F^2$,
\[
\kl{q_t}{\bpi} &= \frac{1}{2}\left[ \log\left( \frac{\det\Sigma}{\det\Sigma(t)} \right) - 2 + \tr\left(\Sigma^{-1}e^{B(t)}\left( I + m(0)m(0)^T \right) e^{B(t)^T}\right)\right]\\
&= \frac{1}{2}\left[ \log\left( \frac{\det\Sigma}{\det\Sigma(t)} \right) - 2 + \left\|\Sigma^{-\frac{1}{2}}e^{B(t)}\left(I+m(0)m(0)^T\right)^{\frac{1}{2}}\right\|_F^2\right].\label{eq:KL_substituted}
\]
From this point onward we will drop the explicit dependence on $t$ for notational brevity.
Note that $B$ has eigendecomposition $B = VDV^{-1}$ where
\[
h &= \frac{g}{2t\sigma^{-1}} \qquad \lambda_+ = h + \sqrt{h^2-1} \qquad \lambda_- = h - \sqrt{h^2-1}\\
V &= t \begin{bmatrix}1 & 0 \\ 0 & \sigma^{-1} \end{bmatrix} \begin{bmatrix} 1 & 1 \\ \lambda_+ & \lambda_- \end{bmatrix} \qquad  D = t\sigma^{-1}\begin{bmatrix} \lambda_+ & 0 \\ 0 & \lambda_- \end{bmatrix}.
\] 
Also note that if $|h|<1$, this decomposition has complex entries, but $e^{B}$ is always a real matrix.

Then the log-determinant term in \cref{eq:KL_substituted} can be written as
\[
\log\left( \frac{\det\Sigma}{\det\Sigma(t)} \right) 
&= \log\frac{\det\Sigma}{\det \Sigma(0)\det e^{2D}} = \log\frac{\sigma^2}{e^{2g}} = \log\frac{\sigma^2}{e^{4ht\sigma^{-1}}} = 2(\log\sigma - 2ht\sigma^{-1}).\label{eq:determinant_term}
\]
By $e^B = Ve^DV^{-1}$, the squared Frobenius norm term in \cref{eq:KL_substituted} is
\[
\|\cdot\|_F^2 &\defas \left\|\Sigma^{-\frac{1}{2}}e^{B(t)}\left(I+m(0)m(0)^T\right)^{\frac{1}{2}}\right\|_F^2\\
&= \frac{\sigma^{-2}\left(e_+-e_-\right)^2}{(\lambda_--\lambda_+)^2}
\left(
(1+\mu^2) + \sigma^2 
+ (1+\mu^2)\left(\frac{\lambda_-e_+-\lambda_+e_-}{e_+-e_-}\right)^2 
+ \sigma^2\left(\frac{\lambda_-e_--\lambda_+e_+}{e_+-e_-}\right)^2 
\right).
\]
where $e_+ = e^{t\sigma^{-1}\lambda_+}, e_- = e^{t\sigma^{-1}\lambda_-}$. Let
\[
\sinh(t,h) = \sinh\left(t\sigma^{-1}\sqrt{h^2-1}\right) \qquad \cosh(t,h) = \cosh\left(t\sigma^{-1}\sqrt{h^2-1}\right),
\]
we further simplify and get
\[
\|\cdot\|_F^2
&= \frac{e^{2th\sigma^{-1}}}{\sigma^2}
\left(
\frac{\sinh(t,h)^2}{h^2-1}(1+\mu^2 + \sigma^2)
+ (1+\mu^2)\left(\frac{h\sinh(t,h)}{\sqrt{h^2-1}} - \cosh(t,h)\right)^2 +\right.\\
&\qquad\qquad\qquad \left.  \sigma^2\left(\frac{h\sinh(t,h)}{\sqrt{h^2-1}} + \cosh(t,h)\right)^2
\right).\label{eq:frobenius_norm_term}
\]
Define $a = t\sigma^{-1}\sqrt{h^2-1}$ and $b = \frac{h}{\sqrt{h^2-1}}$ for $h\neq1$.
Then together with \cref{eq:determinant_term,eq:frobenius_norm_term}, \cref{eq:KL_substituted} can be written as
\[
&\kl{q_t}{\bpi}= \log\sigma - 2ab - 1 + \\
&\frac{e^{2ab}}{2\sigma^2}
\left(
(b^2-1)\sinh(a)^2(1+\mu^2 + \sigma^2)
+ (1+\mu^2)\left(b\sinh(a) - \cosh(a)\right)^2
+  \sigma^2\left(b\sinh(a) + \cosh(a)\right)^2
\right). \label{eq:KL_ab}
\]
Note that if $|h| > 1$, then $a\geq 0$ and $|b| > 1$;
if $|h| < 1$, we can write $a = ia'$ for $a'\geq 0$ and $b = ib'$ for $b' \in (-\infty,-1)\cup(1,\infty)$. We now derive lower bounds for \cref{eq:KL_ab} over $a$ and $b$ under three cases: ($|h|>1,b>1$), ($|h|>1,b<-1$), and ($|h|<1$).

\textbf{Case (}$\bm{|h|>1, b>1}$\textbf{):} Using the identity $\cosh(x)^2-sinh(x)^2=1$ and $\sinh(2x) = 2\sinh(x)\cosh(x)$,
\[
&\frac{e^{2ab}}{2\sigma^2}\left((b^2-1)\sinh(a)^2\sigma^2 + \sigma^2(b\sinh(a)+\cosh(a))^2\right)\\
=& \frac{e^{2ab}}{2}\left((b^2-1)\sinh(a)^2 + (b\sinh(a)+\cosh(a))^2\right) \\
=& \frac{e^{2ab}}{2}\left(2b^2\sinh(a)^2 +1 + b\sinh(2a)\right) \\
\geq& \frac{e^{2ab}}{2} + \frac{b\sinh(2a)}{2} \\
\geq& 2ab,
\]
where the second last line is obtained by noting that $2b^2\sinh(a)^2\geq0$, $e^{2ab}\geq 2ab$, and the last line is obtained by noting that for $|h|>1$ and $b>1$, we have $\sinh(2a)\geq 2a$ and $e^{2ab}\geq1$. Substituting this to \cref{eq:KL_ab},
\[
\kl{q_t}{\bpi} &\geq \log\sigma - 1 + \frac{e^{2ab}}{2\sigma^2}\left( (b^2-1)\sinh(a)^2(1+\mu^2) + (1+\mu^2)(b\sinh(a)-\cosh(a))^2 \right)\\
&\geq \log\sigma  - 1 + \frac{e^{2ab}(1+\mu^2)}{2\sigma^2}
\left(
2b^2\sinh(a)^2 +1 -b\sinh(2a)
\right).\label{eq:case1}
\]
For $|h|>1, b>1$, we know $a\geq0$. When $a\geq0$ and $b>1$, 
\[
\frac{\partial}{\partial a} \left(\log\sigma  - 1 + \frac{e^{2ab}(1+\mu^2)}{2\sigma^2}\left(2b^2\sinh(a)^2 +1 -b\sinh(2a)\right)\right) \geq 0
\]
and increases with $a$. Therefore
\[
\argmin_{a\in\reals_+}\left(\log\sigma  - 1 + \frac{e^{2ab}(1+\mu^2)}{2\sigma^2}
\left(2b^2\sinh(a)^2 +1 -b\sinh(2a)\right)\right) = 0.
\]
We also note that when we set $a=0$, \cref{eq:case1} is constant for all $b>1$. Then by substituting $a=0$ to \cref{eq:case1}, we get
\[
\kl{q_t}{\bpi} \geq &\log\sigma  - 1 + \frac{(1+\mu^2)}{2\sigma^2} \geq \log\frac{1+\mu^2}{4\sigma}.
\]

\textbf{Case (}$\bm{|h|>1, b<-1}$\textbf{):} Note in this case,
$(b^2-1)\sinh(a)^2\sigma^2 \geq 0$, $\sigma^2\left(b\sinh(a)+\cosh(a)\right)^2\geq0$, then we can lower bound \cref{eq:KL_ab} by
\[
\kl{q_t}{\bpi} \geq &\log\sigma - 2ab - 1 +
 \frac{e^{2ab}}{2\sigma^2}
\left(
(b^2-1)\sinh(a)^2(1+\mu^2) + (1+\mu^2)\left(b\sinh(a)-\cosh(a)\right)^2
\right)\\
\geq &\log\sigma - 2ab - 1 +
 \frac{e^{2ab}(1+\mu^2)}{2\sigma^2}
\left(
2b^2\sinh(a)^2 +1 -b\sinh(2a)
\right)\\
 \geq &\log\sigma - 2ab - 1 +
 \frac{e^{2ab}(1+\mu^2)}{2\sigma^2},\label{eq:case2}
\]
where the last line is obtained by noting $2b^2\sinh(a)^2-b\sinh(2a)\geq0$ when $|h|>1$ and $b<-1$. Also when $|h|>1$ and $b<-1$, $ab\leq0$. We then minimize \cref{eq:case2} over $ab \leq 0$, treating $ab$ as a single variable. The stationary point is at
$(ab)^\star = \frac{1}{2}\log\frac{2\sigma^2}{1+\mu^2}$.
Since \cref{eq:case2} is convex in $ab$, the optimum is at $(ab)^\star$ if $\frac{2\sigma^2}{1+\mu^2} \leq 1$, and otherwise is at $(ab) = 0$. Therefore
\[
\kl{q_t}{\bpi}&\geq
\left\{\begin{array}{ll}
 \log\sigma - 1 + \frac{1+\mu^2}{2\sigma^2} & \frac{2\sigma^2}{1+\mu^2} > 1\\
 \log\frac{1+\mu^2}{2\sigma} & \frac{2\sigma^2}{1+\mu^2} \leq 1
\end{array}\right. \\
&\geq \log \frac{1+\mu^2}{4\sigma}.
\]
\textbf{Case (}$\bm{|h|<1}$\textbf{):} Write $a = ia'$ and $b = ib'$ where $a'\geq0$ and $b'\in(-\infty,-1)\cup(1,\infty)$. Using the identities $\sinh(x) = -i\sin(ix)$ and $\cosh(x) = \cos(ix)$, we can write \cref{eq:KL_ab} as
\[
&\kl{q_t}{\bpi} = \log\sigma + 2a'b' - 1 + \\
&\frac{e^{-2a'b'}}{2\sigma^2}
\left(
(b'^2+1)\sin(a)^2(1+\mu^2 + \sigma^2)
+ (1+\mu^2)\left(b'\sin(a') + \cos(a')\right)^2
+  \sigma^2\left(b'\sin(a') - \cos(a')\right)^2
\right)\\
&= \log\sigma + 2a'b' - 1 + \frac{e^{-2a'b'}(1+\mu^2+\sigma^2)}{2\sigma^2}
\left(b'^2+1
+b'f\sin(2a')
-b'^2\cos(2a')
\right),\label{eq:case3}
\]
where $f = \frac{1+\mu^2-\sigma^2}{1+\mu^2+\sigma^2}$. We know
\[
a'^\star = \argmin_{a'\in\reals_+} b'^2+1 + b'f\sin(2a') - b'^2\cos(2a') = \frac{1}{2}\tan^{-1}\left(-\frac{f}{b'}\right) + n\pi,
\]
where $n\in\ints$ such that $\frac{1}{2}\tan^{-1}\left(-\frac{f}{b'}\right) + n\pi \in \reals_+$. Then by $\frac{e^{-2a'b'}(1+\mu^2+\sigma^2)}{2\sigma^2}\geq0$,
\[
\kl{q_t}{\bpi}&\geq \log\sigma + 2a'b' - 1 + \frac{e^{-2a'b'}(1+\mu^2+\sigma^2)}{2\sigma^2}
\left(b'^2+1
+b'f\sin(2a'^\star)
-b'^2\cos(2a'^\star)
\right)\\
&=
\log\sigma + 2a'b' - 1 + \frac{e^{-2a'b'}(1+\mu^2+\sigma^2)}{2\sigma^2}
\left(b'^2+1 - b'^2\sqrt{1+(f/b')^2}\right).
\]
Since $\sqrt{1+x} \leq 1+\frac{1}{2}x$,
\[
\kl{q_t}{\bpi}&\geq
\log\sigma + 2a'b' - 1 + \frac{e^{-2a'b'}(1+\mu^2+\sigma^2)}{2\sigma^2}\left(1 - \frac{1}{2}f^2\right).\label{eq:case3_convex}
\]
The stationary point of \cref{eq:case3_convex} as a function in $a'b'$ is at
\[
(a'b')^\star = -\frac{1}{2}\log\frac{2\sigma^2}{(1+\mu^2+\sigma^2)\left(1-\frac{1}{2}f^2\right)}.
\]
Since \cref{eq:case3_convex} is convex in $a'b'$ and $a'b'\in\reals$, we know the minimum of \cref{eq:case3_convex} is attained at $(a'b')^\star$. Substituting $(a'b')^\star$ back in \cref{eq:case3_convex} and noting $1-\frac{1}{2}f^2 \geq \frac{1}{2}$, we get
\[
\kl{q_t}{\bpi} &\geq \log\frac{(1+\mu^2+\sigma^2)\left(1-\frac{1}{2}f^2\right)}{2\sigma}\\
&\geq \log\frac{1+\mu^2+\sigma^2}{4\sigma} \geq \log\frac{1+\mu^2}{4\sigma}.
\]
\eprfof

\end{document}